\documentclass{article}

\usepackage{latexsym}
\usepackage{graphicx}
\usepackage{mathrsfs}

\newtheorem{theorem}{Theorem}[section]

\newtheorem{corollary}[theorem]{Corollary}

\newtheorem{lemma}[theorem]{Lemma}
\newtheorem{definition}[theorem]{Definition}
\newtheorem{exmp}[theorem]{Example}
\newenvironment{proof}%
\Elproofname\enspace\ignorespaces
\def\Elproofname{{\bf Proof.}}
\endproof

\newcommand{\nd}{\noindent}
\newcommand{\ie}{i.e.}
\newcommand{\eg}{e.g.}
\newcommand{\wrt}{{w.r.t.}}
\newcommand{\ii}[1]{\mbox{$(#1)$}}
\newcommand{\okex}{}

\newcommand{\st}{s.t.}

\newcommand{\ok}{}
\newcommand{\okproof}{}

\newcommand{\okdef}{}

\newcommand{\tuple}[1]{\langle #1 \rangle}
\newcommand{\calB}{{\mathcal B}}
\newcommand{\orderT}{\preceq_{t}}
\newcommand{\orderK}{\preceq_{k}}

\newcommand{\botT}{\mbox{{\tt f}}}
\newcommand{\topT}{\mbox{{\tt t}}}
\newcommand{\botK}{\bot}
\newcommand{\topK}{\top}

\newcommand{\andT}{\wedge}
\newcommand{\orT}{\vee}
\newcommand{\andK}{\otimes}
\newcommand{\orK}{\oplus}

\newcommand{\bigorT}{\bigvee}

\newcommand{\bigorK}{\bigoplus}
\newcommand{\negT}{\neg}
\newcommand{\negK}{\sim}

\newcommand{\lpx}[1]{{\mathcal #1}}
\newcommand{\lp}{\lpx{P}}

\newcommand{\hypox}[1]{#1}

\newcommand{\hypohIx}[1]{\hypox{F}^{#1}}
\newcommand{\hypohI}{\hypohIx{I}}

\newcommand{\lpstarx}[1]{\lpx{#1}^{*}}
\newcommand{\lpstar}{\lpstarx{P}}

\newcommand{\lfp}{\mbox{lfp}}
\newcommand{\gfp}{\mbox{gfp}}

\newcommand{\lfpK}{WF({\lp})}

\newcommand{\IbotT}{\mbox{\tt I}_{\botT}}
\newcommand{\ItopT}{\mbox{\tt I}_{\topT}}
\newcommand{\IbotK}{\mbox{\tt I}_{\botK}}
\newcommand{\ItopK}{\mbox{\tt I}_{\topK}}

\newcommand{\ecopB}{\tilde{\Pi}_{\lp}}

\newcommand{\gammapx}[1]{\Gamma_{#1}}
\newcommand{\gammap}{\gammapx{\lp}}

\newcommand{\phipx}[1]{\Phi_{#1}}
\newcommand{\phip}{\phipx{\lp}}
\newcommand{\psipx}[1]{\Psi_{#1}}
\newcommand{\psip}{\psipx{\lp}}
\newcommand{\psiprimepx}[1]{\Psi'_{#1}}
\newcommand{\psiprimep}{\psiprimepx{\lp}}

\newcommand{\supportpxhx}[2]{s_{#1}(#2)}
\newcommand{\supportpx}[1]{s_{#1}}
\newcommand{\supportp}{\supportpx{\lp}}
\newcommand{\supportph}{\supportpxhx{\lp}{I}}
\newcommand{\supportphIx}[1]{\supportpxhx{\lp}{#1}}

\newcommand{\supportpbotTIx}[1]{\supportpxhx{\lp}{#1}}
\newcommand{\supportphx}[1]{\supportpxhx{\lp}{I}}

\newcommand{\supportClass}[1]{\modcl{\lp \orK #1}}

\newcommand{\supportfunpxhx}[2]{\sigma_{#1}^{#2}}
\newcommand{\supportfunph}{\supportfunpxhx{\lp}{I}}

\newcommand{\supportfunphx}[1]{\supportfunpxhx{\lp}{I}}

\newcommand{\phiprimex}[1]{\Phi'_{#1}}
\newcommand{\phiprimep}{\phiprimex{\lp}}
\newcommand{\iterx}[1]{#1}
\newcommand{\iterJ}{\iterx{J}}
\newcommand{\iterJIx}[1]{\iterJ^{#1}}
\newcommand{\iterJI}{\iterJIx{I}}

\newcommand{\ecop}{\Pi_{\lp}}

\newcommand{\modelscl}{\models_{cl}}

\newcommand{\mod}[1]{mod(#1)}
\newcommand{\modcl}[1]{mod_{cl}(#1)}
\newcommand{\stable}[1]{stable(#1)}

\newif\ifchange
\changetrue      % do print change stuff

\newcommand{\change}[2]  
{\ifchange
     {\bf #1}\mbox{$\to$}{\bf #2}
\else{}
\fi} % example: \change{researchers}{researcher}

\newif\iftodo
\todotrue      % do print todo stuff

\newif\ifcomment
\commenttrue      % do print comment stuff

\title{Epistemic Foundation of Stable Model Semantics \\
{\small \vspace*{1em}
\noindent {\bf To appear in Theory and Practice of Logic Programming (TPLP)}}}

\author{Yann Loyer \\
Laboratoire PRiSM, Universit{\'e} de Versailles Saint Quentin, \\
Versailles, France \\
Yann.Loyer@prism.uvsq.fr \\
\and
Umberto Straccia \\
Istituto di Scienza e Tecnologie dell'Informazione ``A. Faedo", \\
Consiglio Nazionale delle Ricerche\\
Pisa, Italy \\
straccia@isti.cnr.it
}

\begin{document}
\maketitle

\begin{abstract} 
\nd \emph{Stable model semantics} has  become a very popular approach for the management of
negation in logic programming. This approach relies mainly on the closed world assumption to complete the
available knowledge and its formulation has its basis in the so-called
Gelfond-Lifschitz transformation. 

The primary goal of this work is to present an alternative and epistemic-based characterization
of stable model semantics, to the Gelfond-Lifschitz transformation.   In particular, we show
that stable model semantics can be defined entirely as an extension of the Kripke-Kleene
semantics. Indeed, we show that the closed  world assumption can
be seen as an additional  source of `falsehood' to be added \emph{cumulatively} to the
Kripke-Kleene semantics. Our approach is purely algebraic and can abstract from the particular
formalism of choice as it is based on monotone operators (under the knowledge order) over
bilattices only. 

\end{abstract}

%\nd Categories and Subject Descriptors: I.2.3 [{\bf Artificial Intelligence}]:Deduction and
%Theorem Proving - \emph{Logic programming}
%            
%\nd Additional Key Words and Phrases: Bilattices, Fixed-point semantics, Logic
%programs, Stable model semantics, Non-monotonic reasoning

%-----------------------------------
\section{Introduction}
%-----------------------------------

\nd \emph{Stable model semantics}~\cite{Gelfond88,Gelfond91} is probably the
most widely studied and most commonly accepted approach adopted to give meaning to logic programs (with
negation). Informally, it consists in relying on the {\em Closed World Assumption} (CWA) to
\emph{complete} the available knowledge --CWA assumes
that all atoms not entailed by a program are false, see~\cite{Reiter78}, and is motivated by
the fact that explicit representation of negative information in logic programs is not
feasible since the addition of explicit negative information could overwhelm a system.
Defining default rules which allow implicit inference of negated facts from positive
information  encoded in a logic program has been an attractive alternative to the explicit
representation approach.

Stable model semantics defines a whole family of models of (or `answers to') a logic
program and, remarkably, one of these stable models, the {\em minimal} one according to
`knowledge or information ordering', is taken as the favorite (\eg~see~\cite{Denecker98,Denecker01a,Przymusinski90}) and is one-to-one
related with the so-called {\em well-founded semantics}~\cite{vanGelder89,vanGelder91}. 

The original formulation of stable model semantics was classical, two-valued, over the set
of truth-values $\{\botT, \topT\}$.  But,  some programs have no stable
model under this setting. To overcome this problem, Przymusinski~\cite{Przymusinski90a,Przymusinski90b,Przymusinski90} extended the notion of
stable model semantics to allow three-valued, or partial, stable model semantics.  Remarkably,
three-valued logics has also been considered in other approaches for providing semantics to
logic programs, such as  \eg~in \cite{Fitting85,Kunen87} where Clark's completion is
extended to three-valued logics, yielding the well-known 
\emph{Kripke-Kleene semantics} of logic programs. In three-valued semantics, the set of truth
values is $\{\botT, \topT, \botK\}$, where $\botK$ stands for $\emph{unknown}$. Przymusinski
showed that every program has at least a partial stable model and that the well-founded model
is the smallest among them, according to the knowledge ordering. It is then a natural step to
move from a three-valued semantics, allowing the representation of incomplete information, to a
four-valued semantics, allowing the representation of \emph{inconsistency} (denoted $\topK$). 
The resulting semantics is based on the well-known set of truth-values ${\cal FOUR} =
\{\botT, \topT,\botK, \topK\}$, introduced by Belnap~\cite{Belnap77a} to model a kind of
`relevance logic' (there should be some `syntactical' connections between the antecedent and
the consequent of a logical entailment relation $\alpha \models \beta$, --see
also~\cite{Anderson75,Dunn76,Dunn86,Levesque84a,Levesque88}. This process of enlarging the set
of  truth-values culminated with Fitting's progressive
work~\cite{Fitting93,Fitting02,Fitting85,Fitting91,Fitting92} on giving meaning to logic
programs by relying on \emph{bilattices}~\cite{Ginsberg88a}. Bilattices, where $\cal FOUR$ is
the simplest non-trivial one, play an important role in logic programming, and in knowledge
representation in general. Indeed, Arieli and Avron show~\cite{Arieli96,Arieli98} that the 
use  of four values is preferable to the use of three values even for 
tasks that can in principle be handled using only three values. Moreover, Fitting explains 
clearly~\cite{Fitting91} why  $\mathcal{FOUR}$ can be thought of as the `home' of classical
logic programming. Interestingly, the algebraic work of Fitting's  fixed-point characterisation
of stable model semantics on bilattices~\cite{Fitting93,Fitting02} has been the starting point of the
work carried out by Denecker, Marek and
Truszczy{\'n}ski~\cite{Denecker99,Denecker02,Denecker03}, who extended Fitting's work to a more
abstract context of fixed-points operators on lattices, by relying on \emph{interval}
bilattices (these bilattices are obtained in a standard way as a product of a lattice --see, for instance~\cite{Fitting93,Fitting92}). Denecker, Marek and Truszczy{\'n}ski~\cite{Denecker99,Denecker03} showed
 interesting connections between (two-valued and four-valued)
Kripke-Kleene~\cite{Fitting85}, well-founded and stable model semantics, as well as to Moore's
autoepistemic logic~\cite{Moore84} and Reiter's default logic~\cite{Reiter80}. Other
well-established applications of bilattices and/or 
Kripke-Kleene, well-founded and stable model semantics to give semantics to logic programs can
be found in the context of reasoning  under paraconsistency and uncertainty
(see, \eg%
~\cite{Alcantara02,Arieli02,BlairH89,Damasio98,Damasio01,Loyer02d,Loyer02b,Loyer03c,Loyer03d,LukasiewiczT01,Ng91}).
Technically, classical two-valued stable models of logic programs are defined in terms of 
fixed-points of the so-called Gelfond-Lifschitz operator, $GL(I)$, for a two-valued
interpretation $I$. This operator has been generalized to bilattices by
Fitting~\cite{Fitting93}, by means of the $\psiprimep(I)$ operator, where this time $I$ is
an interpretation over bilattices. Informally, the main principle of these
operators is based on the \emph{separation of the role of positive and negative information}.
That is, given a two-valued interpretation $I$, $GL(I)$ is obtained by
first evaluating negative literals in a logic program $\lp$ by means of $I$, determining the
\emph{reduct} $\lp^{I}$ of $\lp$, and then, as $\lp^{I}$ is now a positive program, to compute
the minimal Herbrand model of $\lp^{I}$ by means of the usual Van Emden-Kowalski's immediate
consequence operator $T_\lp$~\cite{vanEmden76,Lloyd87}. The computation of $\psiprimep(I)$ for bilattices is similar. As a consequence, this separation avoids the natural management of
classical negation (\ie~the  evaluation of a negative literal $\negT A$ is given by the negation
of the evaluation of $A$), which is a major feature of the Kripke-Kleene semantics~\cite{Fitting85,Fitting91} of logic programs with negation.

The primary goal of this study is to show, in the quite general setting of bilattices as space
of truth-values, that this separation of positive and negative information is nor necessary
nor is any program transformation required to characterize  stable model
semantics epistemologically. Another motivation is to evidence the role of CWA as a discriminating factor
between the most commonly accepted semantics of logic programs. We show that the only 
difference between Kripke-Kleene, well-founded and stable model semantics 
is the amount of knowledge taken from CWA that they integrate.
%Indeed, we show that 
%the well-founded and stable model semantics can be defined as epistemic
%extensions of the  Kripke-Kleene  semantics. 
We view CWA, informally as an additional source
of information to be used for information completion, or more precisely, as a carrier of
falsehood, to be considered cumulatively to Kripke-Kleene semantics. This allows us to view
stable model semantics from a different, not yet investigated 
perspective.  Roughly speaking, in Kripke-Kleene semantics, CWA is used  to consider  
only those atoms that do not appear in head of any rule as false (and that can obviously not be inferred as true),
while  the well-founded and stable model semantics integrate more  CWA-provided knowledge. To identify this knowledge, we introduce the notion of \emph{support}. 
This is a generalization of the notion of \emph{greatest unfounded set}~\cite{vanGelder91} (which
determines the atoms that can be assumed to be false) to the bilattice context. It 
determines in a principled way the amount of falsehood provided by CWA that can "safely" be assumed. More precisely, as we are considering a many-valued truth space, it provides the degree of falseness that can "safely" be assumed for each atom. We then show how  the support can be used to complete the Kripke-Kleene semantics in order to obtain the well-founded and  stable model semantics over bilattices.
In particular, we show that the well-founded semantics is the least informative model
in the set of models containing their own support, while a model is a stable model if and only 
if it is deductively closed under support completion, \ie~it contains exactly the knowledge that can be inferred by activating the rules over the support.  
We thus show an alternative characterisation of the stable model semantics to the well-known, widely applied and long studied technique based on the separation of positive and negative information in the Gelfond-Lifschitz transformation, 
by reverting to the classical interpretation of negation. 
While the Gelfond-Lifschitz transformation treats negation-as-failure in a special way and unlike other connectives, our approach is an attempt to relate the semantics of logic programs to a standard model-theoretic account of rules. We show that logic programs can be analyzed using standard logical means such as the notion of interpretation and information ordering, \ie~knowledge
ordering. Therefore, in principle, our approach  does not depend on the presence of any specific
connective, such as negation-as-failure, nor on any specific rule syntax (the work of Herre and Wagner~\cite{Herre97}, is in this direction, even if it differs slightly from the usual stable model semantics~\cite{Gelfond91} and the semantics is given in the context of the classical, two-valued truth-space). Due to the
generality and the purely algebraic nature of our results, as just monotone operators over
bilattices are postulated, the epistemic characterisation of stable models given in this study
can also be applied in other contexts (\eg~uncertainty and/or paraconsistency in logic programming, and nonmonotonic logics such as default and autoepistemic logics).

The rest of the paper is organized as follows. In order to make the paper self-contained,
in the next section, we will provide definitions and properties of bilattices and logic
programs. Section~\ref{ecopdef} is the main part of this work, where we present our
characterisation of the stable model semantics, while Section~\ref{concl} concludes.

%-----------------------------------
\section{Preliminaries}
%-----------------------------------

%-----------------------------------
\subsection{Lattices}
%-----------------------------------

\nd A \emph{lattice} is a partially ordered set $\tuple{L,\preceq}$ such that every two
element set $\{x,y\} \subseteq L$ has a \emph{least upper bound}, $lub_{\preceq}(x,y)$
(called the \emph{join} of $x$ and $y$), and a \emph{greatest lower bound},
$glb_{\preceq}(x,y)$ (called the \emph{meet} of $x$ and $y$). For ease, we will write $x \prec
y$ if $x \preceq y$ and $x \neq y$.  A lattice $\tuple{L,\preceq}$ is \emph{complete} if every
subset of $L$ has both least upper and
greatest lower bounds. Consequently, a complete lattice has a least element, $\bot$, and a
greatest element $\top$.
For ease, throughout the paper, given a complete lattice $\tuple{L,\preceq}$ and a subset of
elements $S \subseteq L$, with \emph{$\preceq$-least} and \emph{$\preceq$-greatest} we will
always mean $glb_{\preceq}(S)$ and $lub_{\preceq}(S)$, respectively. With $\min_{\preceq}(S)$
we denote the set of minimal elements in $S$ \wrt~$\preceq$, 
\ie~$\min_{\preceq}(S) = \{x \in S \colon  \not \exists y \in S~\st~y \prec x\}$. Note that
while $glb_{\preceq}(S)$ is unique, $|\min_{\preceq}(S)| > 1$ may hold. If
$\min_{\preceq}(S)$ is a singleton $\{x\}$, for convenience we may also write $x =
\min_{\preceq}(S)$ in place of $\{x\} = \min_{\preceq}(S)$.
An \emph{operator} on a lattice $\tuple{L,\preceq}$ is a function from $L$ to $L$, 
$f\colon L \to L$. An operator $f$ on $L$ is \emph{monotone}, if for every pair of elements $x,y
\in L$, $x \preceq y$ implies $f(x)\preceq f(y)$, while $f$ is \emph{antitone} if
$x\preceq y$ implies $f(y) \preceq f(x)$. A
\emph{fixed-point} of $f$ is an element $x \in L$ such that $f(x)=x$.

The basic tool for studying fixed-points of operators on lattices is the
well-known Knaster-Tarski theorem~\cite{Tarski55}.

\begin{theorem}[Knaster-Tarski fixed-point theorem~\cite{Tarski55}] \label{thm.kt}
Let $f$ be a monotone operator on a complete lattice $\tuple{L,\preceq}$. 
Then $f$ has a fixed-point, the set of fixed-points of $f$ is a complete lattice and, thus,
$f$ has a \emph{$\preceq$-least} and a \emph{$\preceq$-greatest} fixed-point.
The \emph{$\preceq$-least} (respectively, \emph{$\preceq$-greatest}) fixed-point can be
obtained by iterating $f$ over $\bot$ (respectively, $\top$), \ie~is the limit of the 
non-decreasing (respectively, non-increasing) sequence $x_{0},
\ldots, x_{i}, x_{i+1}, \ldots, x_{\lambda}, \ldots$, where for a successor ordinal $i \geq 0$,
\begin{eqnarray*}
x_{0} & = & \bot, \\
x_{i+1} & = & f(x_{i})
\end{eqnarray*}

\nd (respectively, $x_{0} = \top$), while for a limit ordinal $\lambda$, 

\begin{equation} \label{eq.lattice}
x_{\lambda} = lub_{\preceq}\{x_{i}\colon i < \lambda \} \
(respectively, x_{\lambda} = glb_{\preceq}\{x_{i}\colon i < \lambda \})  \ . 
\end{equation}
\ok 
\end{theorem}

\nd We denote the $\preceq$-least and the $\preceq$-greatest fixed-point by $\lfp_{\preceq}(f)$
and $\gfp_{\preceq}(f)$, respectively. 

Throughout the paper, we will frequently define monotone operators, whose sets of fixed-points
define  certain classes of models of a logic program. As a consequence, please note that this
also means that a least model {\em always} exists for such classes. Additionally, for ease,
for the monotone operators defined in this study, we will specify the initial condition $x_{0}$
and the next iteration step $x_{i+1}$ only, while Equation~(\ref{eq.lattice}) is always considered as
implicit. To prove that a property holds for a limit ordinal of an iterated sequence, \ie~for transfinite induction, one usually relies on a routine least upper bound (or greatest lower bound) argument and on the Knaster-Tarski theorem. Therefore that case  will be considered only in the proof of Theorem~\ref{thm.supportcomp}, while the reasoning is similar for all the other proofs and, thus, will be omitted.
 
%-----------------------------------
\subsection{Bilattices}
%-----------------------------------

\nd The simplest non-trivial bilattice, called $\mathcal{FOUR}$, was defined by Belnap~\cite{Belnap77a} (see
also~\cite{Arieli98,Avron96,Ginsberg88a}), 
who introduced a logic intended to deal with incomplete and/or inconsistent information.
$\mathcal{FOUR}$ already illustrates many of the basic properties of bilattices.
Essentially, it extends the classical truth set $\{\botT,\topT\}$ to its
power set $\{\{\botT\},\{\topT\}, \emptyset, \{\botT, \topT\}\}$, where we can think that each
set indicates the amount of information we have in terms of truth: so, $\{\botT\}$ stands for
\emph{false}, $\{\topT\}$ for \emph{true} and, quite naturally, $\emptyset$ for lack of
information or \emph{unknown}, and $\{\botT, \topT\}$ for \emph{inconsistent} information (for
ease, we use $\botT$ for $\{\botT\}$, $\topT$ for $\{\topT\}$, $\botK$ for $\emptyset$ and
$\topK$ for $\{\botT, \topT\}$).  The set of truth values $\{\botT,\topT, \botK, \topK\}$ has
two quite intuitive and natural `orthogonal' orderings, $\orderK$ and $\orderT$ (see
Figure~\ref{four}), each giving to $\mathcal{FOUR}$ the structure of a complete lattice. 
\begin{figure}[h]
\begin{center}
\setlength{\unitlength}{0.025cm}
\begin{picture}(150,150)
\thinlines
\put(20,20){\vector(1,0){130}}
\put(20,20){\vector(0,1){130}}
\put(135,5){$\orderT$}
\put(0,135){$\orderK$}

\put(85,40){\circle*{5}}
\put(85,120){\circle*{5}}
\put(45,80){\circle*{5}}
\put(125,80){\circle*{5}}

\put(80,25){$\botK$}
\put(80,125){$\topK$}
\put(32,75){\botT}
\put(130,75){$\topT$}

\put(85,40){\line(1,1){40}}
\put(85,120){\line(-1,-1){40}}
\put(45,80){\line(1,-1){40}}
\put(125,80){\line(-1,1){40}}

\end{picture}
\end{center}
\caption{The logic ${\mathcal{FOUR}}$.} \label{four}
\end{figure}
One is the so-called
\emph{knowledge ordering}, denoted $\orderK$, and is based on the subset relation, that is, if
$x \subseteq y$ then $y$ represents `more information' than $x$ (\eg~$\botK = \emptyset
\subseteq \{\topT\} = \topT$, \ie~$\botK  \orderK \topT$). 
The other ordering is the so-called \emph{truth ordering}, denoted $\orderT$. Here $x \orderT y$
means that $x$ is `at least as false as $y$, 
and $y$ is at least as true as $x$', \ie~$x
\cap \{\topT\} \subseteq y \cap \{\topT\}$ and $y \cap \{\botT\} \subseteq x \cap \{\botT\}$
(\eg~$\botK \orderT \topT$).

The general notion of bilattice used in this paper is defined as
follows~\cite{Fitting02,Ginsberg88a}. A \emph{bilattice} is a structure $\tuple{\calB, \orderT,
\orderK}$  where $\calB$ is a non-empty set and $\orderT$ and $\orderK$ are both partial
orderings giving $\calB$ the structure of a \emph{complete lattice} with a top and bottom
element.
 {\em Meet and join under $\orderT$}, denoted
$\andT$ and $\orT$, correspond to extensions of classical conjunction and disjunction. On the other hand,
{\em meet and join under $\orderK$} are denoted $\andK$ and $\orK$. $x \andK y$ corresponds to
the maximal information $x$ and $y$ can agree on, while  $x \orK y$ simply combines the
information represented by $x$ with that represented by $y$. {\em Top and bottom under
$\orderT$} are denoted $\topT$ and $\botT$, and {\em top and 
bottom under $\orderK$} are denoted $\topK$ and $\botK$, respectively. 
We will assume that bilattices are
\emph{infinitary distributive bilattices} in which all distributive laws 
connecting $\andT, \orT, \andK$ and $\orK$ hold. We also assume that every bilattice satisfies
the \emph{infinitary interlacing conditions}, \ie~each of the lattice operations 
$\andT, \orT,\andK$ and $\orK$ is monotone \wrt~both orderings. 
An example of interlacing condition is: $x \orderT y$ and $x' \orderT y'$ implies $x \andK x'
\orderT y \andK y'$. Finally, we assume that each bilattice has a \emph{negation},
\ie~an operator $\negT$ that reverses the $\orderT$ ordering, leaves unchanged the $\orderK$
ordering, and verifies $\negT \negT x = x$~\footnote{The dual operation to negation is
\emph{conflation}, \ie~an operator $\negK$ that reverses the
$\orderK$ ordering, leaves unchanged the $\orderT$ ordering, and 
$\negK \negK x = x$. If a bilattice has both, they
\emph{commute} if $\negK \negT x = \negT \negK x$ for all $x$. 
We will not deal with conflation in this paper.}.

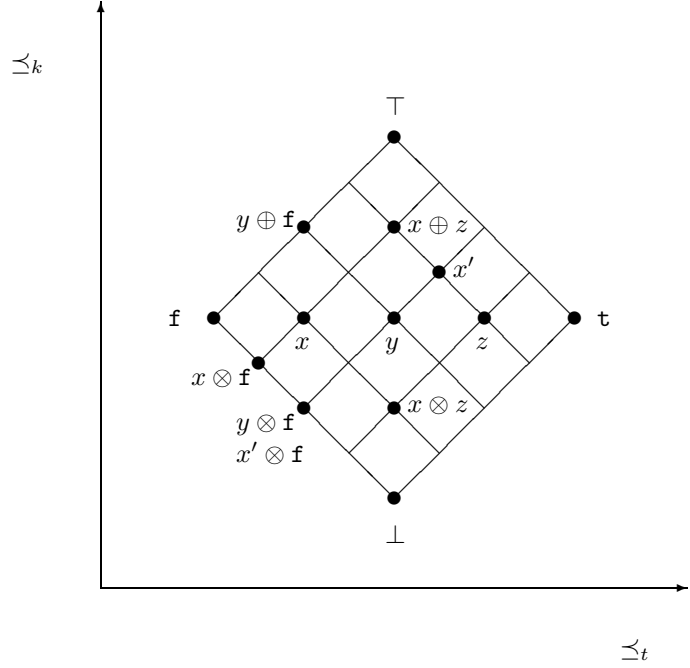
\begin{figure}[h]
\begin{center}
\setlength{\unitlength}{0.06cm}
\begin{picture}(150,150)
\thinlines
\put(20,20){\vector(1,0){130}}
\put(20,20){\vector(0,1){130}}
\put(135,5){$\orderT$}
\put(0,135){$\orderK$}

\put(85,40){\circle*{3}}
\put(85,120){\circle*{3}}
\put(45,80){\circle*{3}}
\put(125,80){\circle*{3}}

\put(83,30){$\botK$}
\put(83,125){$\topK$}
\put(35,78){\botT}
\put(130,78){$\topT$}

\put(85,40){\line(1,1){40}}
\put(85,120){\line(-1,-1){40}}
\put(45,80){\line(1,-1){40}}
\put(125,80){\line(-1,1){40}}

\put(75,50){\line(1,1){40}}
\put(65,60){\line(1,1){40}}
\put(55,70){\line(1,1){40}}

\put(115,70){\line(-1,1){40}}
\put(105,60){\line(-1,1){40}}
\put(95,50){\line(-1,1){40}}

%\put(85,40){\line(0,1){80}}

\put(65,80){\circle*{3}}
\put(85,60){\circle*{3}}
\put(85,100){\circle*{3}}
\put(105,80){\circle*{3}}
\put(85,80){\circle*{3}}
\put(55,70){\circle*{3}}
\put(65,60){\circle*{3}}
\put(65,100){\circle*{3}}
\put(95,90){\circle*{3}}

\put(63,73){$x$}
\put(103,73){$z$}
\put(83,73){$y$}
\put(88,59){$x\andK z$}
\put(88,99){$x\orK z$}
\put(40,65){$x\andK \botT$}
\put(50,55){$y\andK \botT$}
\put(50,48){$x'\andK \botT$}
\put(50,100){$y\orK \botT$}
\put(98,89){$x'$}
%\put(100,45){$y \orderT x', y \orderK x'$}

%\put(97,47){$\alpha\andK\topT$}
%\put(100,110){$\alpha\orK\topT$}

%\put(83,60){\linethickness{0.5mm} \line(0,1){40}}

%\put(63,79){\linethickness{2.4mm} \line(1,1){31}}
%\put(63,80){\linethickness{2.4mm} \line(1,-1){31}}
%\put(94,110){\linethickness{2.4mm} \line(1,-1){31}}
%\put(93,50){\linethickness{2.4mm} \line(1,1){31}}

%\put(55,70){\circle*{3}}
%\put(45,40){$\supportphIx{I_{1}}$}
%\put(25,60){$\supportphIx{I_{2}}$}
%
%\put(95,70){\circle*{3}}
%\put(85,100){\circle*{3}}
%\put(100,67){$I_{1}$}
%\put(90,97){$I_{2}$}

\end{picture}
\end{center}
\caption{Some points mentioned in Lemmas~\ref{prop.y1}--\ref{prop.sms4}.}
\label{fig.lembilattices}
\end{figure}

Below, we give some properties of bilattices 
that will be used in this study. Figure~\ref{fig.lembilattices} illustrates intuitively some of
the following lemmas.

\begin{lemma}[\cite{Fitting93}]\label{prop.y1}
\mbox{}
\begin{enumerate}
\item If $x \orderT y \orderT z$ then $x \andK z \orderK y$ and $y \orderK x \orK z$;
\item If $x \orderK y \orderK z$ then $x \andT z \orderT y$ and $y \orderT x \orT z$. \ok
\end{enumerate}
\end{lemma}

\begin{lemma}\label{prop.p1}
If $x \orderT y$ then $x \orderT x \andK y \orderT y$ and $x \orderT 
x \orK y \orderT y$. \ok
\end{lemma}
\begin{proof}
Straightforward using the interlacing conditions. \okproof
\end{proof}

\begin{lemma}\label{prop.p9}\mbox{ \ }
\begin{enumerate}
\item If $x 
\orderT y$ then $\botT \andK x \orderT y$;
\item If $x \orderK y$ 
then $\botT \andK y \orderT x$. \ok
\end{enumerate}
\end{lemma}
\begin{proof}
If $x \orderT y$ then from $\botT \orderT x$ 
and by Lemma~\ref{prop.p1}, 
$\botT \orderT \botT \andK x 
\orderT x \orderT y$.
If $x \orderK y$ then, from $\botT \orderT x$, we have $\botT \andK y \orderT x \andK y = x$.
\okproof
\end{proof}

\begin{lemma}\label{prop.sms2}
If $x \orK z \orderT y$ then $z \orderK y \orK \botT$. \ok
\end{lemma}
\begin{proof}
By Lemma~\ref{prop.y1}, $\botT \orderT x \orK z \orderT y$ 
implies $z \orderK x \orK z \orderK y \orK \botT$. \okproof
\end{proof}

\begin{lemma}\label{prop.sms3}
If $\botT \andK y \orderK x \orderK \botT \orK y$ then $x \orderT y$. \ok
\end{lemma}
\begin{proof}
By Lemma~\ref{prop.y1}, $\botT \andK y \orderK x \orderK \botT 
\orK y$ implies 
$x\orderT (\botT \andK y) \orT (\botT \orK y)$. 
Therefore, 
$x \orderT (\botT \andK y) \orK ((\botT \andK y) \orT y) 
$ and, thus, 
$x \orderT (\botT \andK y) \orK y = y$. \okproof
\end{proof}

\begin{lemma}\label{prop.sms4}
If $x \orderK y$ and $x \orderT y$ then $x \andK \botT = y \andK \botT$. \ok
\end{lemma}
\begin{proof}
By Lemma~\ref{prop.p9}, $\botT \andK y \orderT x$ and, thus, 
$\botT \andK y \orderT x \andK \botT$ follows. From $x \orderT y$, 
$\botT \andK x \orderT y \andK \botT$ holds. Therefore,  $x \andK 
\botT = y \andK \botT$. \okproof
\end{proof}

\subsubsection{Bilattice construction} \label{b.constr}
Bilattices come up in natural ways. There are two
general, but different, construction methods, to build a bilattice from a
lattice which are widely used. We only outline them here in order to give an idea of their
application (see also~\cite{Fitting93,Ginsberg88a}).

The first bilattice construction method was proposed by~\cite{Ginsberg88a}. Suppose we 
have two complete distributive lattices $\tuple{L_{1},\preceq_{1}}$ and
$\tuple{L_{2},\preceq_{2}}$.
Think of $L_{1}$ as a lattice of values we use when we measure the
degree of belief, while think of $L_{2}$ as the lattice we use
when we measure the degree of doubt. Now, we define the structure $L_{1} \odot L_{2}$ as
follows. The structure is $\tuple{L_{1} \times L_{2}, \orderT, \orderK}$, where

\begin{itemize}
\item  $\tuple{x_{1}, x_{2}} \orderT \tuple{y_{1}, y_{2}}$ if $x_{1} \preceq_{1} y_{1}$ and
$y_{2} \preceq_{2} x_{2}$;
\item  $\tuple{x_{1}, x_{2}} \orderK \tuple{y_{1}, y_{2}}$ if $x_{1} \preceq_{1} y_{1}$ and
$x_{2} \preceq_{2} y_{2}$.
\end{itemize}

\nd In  $L_{1} \odot L_{2}$ the idea is: knowledge goes up if both degree of belief and degree 
of doubt go up; truth goes up if the degree of belief goes up, while
the degree of doubt goes down. It can easily be verified that $L_{1} \odot
L_{2}$ is a bilattice. 
Furthermore, if $L_{1} = L_{2} = L$, \ie~we are measuring belief and doubt in the same way
(\eg~$L = \{\botT,\topT\}$), then negation can be defined as $\negT \tuple{x,y} = \tuple{y,x}$,
\ie~negation switches the roles of belief and doubt. Applications of this
method can be found, for instance, 
in~\cite{Alcantara02,Ginsberg88a,Herre97}.

The second construction method has been sketched in~\cite{Ginsberg88a} and addressed in
more detail in~\cite{Fitting92}, and is probably the more used one. Suppose we have a  complete
distributive lattice of truth values $\tuple{L,\preceq}$. Think of these values as the `real'
values in which we are interested, but due to lack of knowledge we are able just to `approximate' the
exact values. Rather than considering a pair $\tuple{x,y} \in L \times L$ as indicator
for degree of belief and doubt, $\tuple{x,y}$ is interpreted as the set of elements $z \in L$
such that $ x \preceq z \preceq y$. That is, a pair $\tuple{x,y}$ is interpreted as an
\emph{interval}. An interval $\tuple{x,y}$ may be seen as an approximation of an exact value.
For instance, in reasoning under uncertainty~(see, \eg~\cite{Loyer02b,Loyer03c,Loyer03d}), $L$ is the
unit interval $[0,1]$ with standard ordering, $L \times L$ is interpreted as the set of (closed)
intervals in $[0,1]$, and the pair $\tuple{x,y}$ is interpreted as a lower and an upper bound of
the exact value of the certainty value. A similar interpretation is given
in~\cite{Denecker99,Denecker02,Denecker03}, but this time $L$
is the set of two-valued interpretations, and a pair $\tuple{J^{-}_{I},J^{+}_{I}} \in L \times L$
is interpreted as a lower and upper bound approximation of the application of a monotone
(immediate consequence) operator $O\colon L \to L$ to an interpretation $I$.

Formally, given the lattice $\tuple{L,\preceq}$, the \emph{bilattice of intervals} is $\tuple{L
\times L, \orderT,
\orderK}$, where:
\begin{itemize}
\item  $\tuple{x_{1}, x_{2}} \orderT \tuple{y_{1}, y_{2}}$ if $x_{1} \preceq y_{1}$ and
$x_{2} \preceq y_{2}$;
\item  $\tuple{x_{1}, x_{2}} \orderK \tuple{y_{1}, y_{2}}$ if $x_{1} \preceq y_{1}$ and
$y_{2} \preceq x_{2}$.
\end{itemize}
 
\nd The intuition of these orders is that truth
increases if the interval contains greater values, whereas the
knowledge increases when the interval becomes more precise. Negation can be defined as $\negT
\tuple{x,y} = \tuple{\negT y, \negT x}$, where $\negT$ is a negation operator on $L$. Note that,
if $L = \{\botT, \topT\}$, and if we assign $\botT = \tuple{\botT, \botT}$, $\topT =
\tuple{\topT, \topT}$, $\botK = \tuple{\botT, \topT}$ and $\topK
= \tuple{\topT, \botT}$, then we obtain a structure that is isomorphic to the bilattice $\cal FOUR$.

%-----------------------------------
\subsection{Logic programs, interpretations, models and program knowledge completions}
%-----------------------------------

\nd We recall here the definitions given in~\cite{Fitting93}. This setting is as general as
possible, so that the results proved in this paper will be widely applicable. 

Classical logic programming has the set $\{\botT, \topT\}$ as its truth space, but as stated by
Fitting~\cite{Fitting93}, ``$\cal FOUR$ can be thought as the `home' of ordinary logic
programming and its natural extension is to bilattices other than $\cal FOUR$: the more general
the setting the more general the results". We will also consider bilattices as the truth space of
logic programs.

\subsubsection{Logic programs} Consider an alphabet of predicate symbols, of constants, of function symbols and variable symbols. A \emph{term}, $t$, is either a variable $x$, a constant $c$ or of the form $f(t_{1}, \ldots, t_{n})$, where $f$ is an $n$-ary function symbol and all $t_{i}$ are terms. An
\emph{atom}, $A$, is of the form $p(t_{1}, \ldots, t_{n})$, where $p$ is an $n$-ary predicate
symbol and all $t_{i}$ are terms. A literal, $l$, is of the form $A$ or $\neg A$, where $A$ is an atom.
A \emph{formula}, $\varphi$, is an expression built up from  the  literals  and the
members of a bilattice $\calB$ using $\andT, \orT, \andK,\orK,\exists$ and $\forall$. 
Note that members of the bilattice may appear in a formula, 
\eg~in $\mathcal{FOUR}$, $(p \andT q) \orK (r \andK \botT)$ is a formula.
A \emph{rule} is of the form  $p(x_{1}, \ldots, x_{n}) \leftarrow  \varphi(x_{1}, \ldots, 
x_{n})$, where $p$ is an $n$-ary predicate symbol and all $x_{i}$ are variables. The
atom $p(x_{1}, \ldots, x_{n})$ is called the \emph{head}, and the formula
$\varphi(x_{1}, \ldots, x_{n})$ is called the \emph{body}. It is assumed that the free
variables of the body are among $x_{1}, \ldots, x_{n}$. Free variables are thought of as
universally quantified. A \emph{logic program}, denoted with $\lp$, is a finite set of
rules. The \emph{Herbrand universe} of $\lp$ is the set of \emph{ground} (variable-free)
terms that can be built from the constants and function symbols occurring in $\lp$, while
the \emph{Herbrand base} of $\lp$  (denoted $B_\lp$) is the set of ground atoms over the Herbrand
universe. 

\begin{definition}[$\lpstar$] \label{lpstar}
Given a logic program $\lp$, the associated set $\lpstar$ is constructed as follows;
\begin{enumerate}
\item put in $\lpstar$ all ground instances of members of $\lp$ (over the  Herbrand base);
\item if a ground atom $A$ is not head of any rule in $\lpstar$, then add the rule $A \leftarrow
\botT$ to $\lpstar$. Note that it is a standard practice in logic programming to consider such
atoms as \emph{false}. We incorporate this by explicitly adding $A \leftarrow \botT$ to $\lpstar$;
\item replace several ground rules in $\lpstar$ having same head, $A \leftarrow \varphi_{1}$, $A
\leftarrow \varphi_{2}$, \ldots with $A \leftarrow \varphi_{1} \orT \varphi_{2} \orT \ldots$.
As there could be infinitely many grounded rules with same head, we may end with a countable
disjunction, but the semantics behavior is unproblematic. \okdef
\end{enumerate}

\end{definition}

\nd Note that in $\lpstar$, each ground atom appears in the head of \emph{exactly one} rule.

\subsubsection{Interpretations} Let $\tuple{\calB, \orderT, \orderK}$ be a bilattice. By 
\emph{interpretation of a logic program} on the  bilattice we mean a mapping $I$ from 
ground  atoms to members of $\calB$. An interpretation $I$ is extended from atoms to formulae as follows:
\begin{enumerate}
\item for $b \in \calB$, $I(b) = b$; 
\item for formulae $\varphi$ and $\varphi'$, $I(\varphi \andT \varphi') = I(\varphi) 
\andT I(\varphi')$, and similarly for $\orT, \andK,\orK$ and $\negT$; and
\item $I(\exists x \varphi(x)) = \bigorT \{I(\varphi(t))\colon t \mbox{ ground term} \}$, and
similarly for universal quantification~\footnote{The bilattice is complete \wrt~$\orderT$, so
existential and universal quantification are well-defined.}.
\end{enumerate}

\nd The family of all interpretations is denoted by ${\mathcal I}(\calB)$.
The truth and knowledge orderings are extended from $\calB$ to ${\mathcal I}(\calB)$ as
follows:

\begin{itemize}
\item $I_{1} \orderT I_{2}$ iff $I_{1}(A)  \orderT I_{2}(A)$, for 
every ground atom $A$; and
\item $I_{1} \orderK I_{2}$ iff $I_{1}(A)  \orderK I_{2}(A)$, for 
every ground atom $A$.
\end{itemize}

\nd Given two interpretations $I,J$, we define $(I \andT J)(\varphi) =
I(\varphi) \andT J(\varphi)$, and similarly for the other
operations.  With $\IbotT$ and $\ItopT$ we denote the
bottom and top interpretations under $\orderT$ (they map any atom
into $\botT$ and $\topT$, respectively).  With $\IbotK$ and
$\ItopK$ we denote the bottom and top interpretations under $\orderK$
(they map any atom into $\botK$ and $\topK$, respectively).
It is easy to see that the space of interpretations $\tuple{{\mathcal
I}(\calB), \orderT, \orderK}$ is an infinitary interlaced and distributive bilattice as well.

\subsubsection{Classical setting}
Note that in a \emph{classical logic program} the body is a conjunction of literals.
Therefore, if $A \leftarrow \varphi \in \lpstar$, then $\varphi = \varphi_1 \orT \ldots
\orT \varphi_n$ and $\varphi_i= L_{i_1} \andT \ldots \andT L_{i_n}$. Furthermore, a
\emph{classical total interpretation} is an interpretation over $\cal FOUR$ such that an atom
is mapped into either $\botT$ or $\topT$. A \emph{partial classical interpretation} is a
classical interpretation where the truth of some atom may be left unspecified. This is the same
as saying that the interpretation maps all atoms into either $\botT,\topT$ or $\botK$. For a set
of literals $X$, with $\neg.X$ we indicate the set $\{\neg L\colon L \in X\}$, where for any
atom $A$, $\neg \neg A$ is replaced with $A$. Then, a classical interpretation (total or
partial) can also be represented as a consistent set of literals, \ie~$I \subseteq B_\lp \cup
\neg.B_\lp$  and for all atoms $A$, $\{A,\neg A\} \not\subseteq I$. Of course, the opposite is
also true, \ie~a consistent set of literals can straightforwardly be turned into an
interpretation over $\cal FOUR$.

\subsubsection{Models} An interpretation $I$ is a \emph{model} of a logic program $\lp$, denoted
by $I \models \lp$, if and only if for each rule $A \leftarrow\varphi$ in $\lpstar$,
$I(\varphi) \orderT I(A)$. With $\mod{\lp}$ we identify the set of models of $\lp$. 

From all models of a logic program $\lp$, Fitting~\cite{Fitting93,Fitting02} identifies a
subset, which obeys  the so-called \emph{Clark-completion} procedure~\cite{Clark78}. 
Essentially, we replace in $\lpstar$ each occurrence of $\leftarrow$ with $\leftrightarrow$: 
an interpretation $I$ is a \emph{Clark-completion model}, \emph{cl-model} for short, of a logic
program $\lp$, denoted by $I \modelscl \lp$, if and only if for each rule $A
\leftarrow\varphi$ in $\lpstar$, $I(A) = I(\varphi)$. With $\modcl{\lp}$ we identify the set
of cl-models of $\lp$. Of course $\modcl{\lp} \subseteq \mod{\lp}$ holds.

\begin{exmp} \label{ex.models}
Consider the following logic program

\[
\lp = \{(A \leftarrow \negT A), (A \leftarrow \alpha) \} \ ,
\]

\nd where $\alpha$ is a value of a bilattice such that $\alpha \orderT \negT \alpha$ and $A$
is a ground atom. Then $\lpstar$ is

\[
\lpstar = \{A \leftarrow \negT A \orT \alpha \} \ .
\]
  
\nd Consider Figure~\ref{fig.1}.
\begin{figure}[h]
\begin{center}
\setlength{\unitlength}{0.07cm}
\begin{picture}(135,135)
\thinlines
\put(20,10){\vector(1,0){120}}
\put(20,10){\vector(0,1){120}}
\put(130,0){$\orderT$}
\put(10,120){$\orderK$}

\put(85,40){\circle*{3}}
\put(85,120){\circle*{3}}
\put(45,80){\circle*{3}}
\put(125,80){\circle*{3}}

\put(83,30){$\botK$}
\put(83,125){$\topK$}
\put(35,79){\botT}
\put(130,79){$\topT$}

\put(85,40){\line(1,1){40}}
\put(85,120){\line(-1,-1){40}}
\put(45,80){\line(1,-1){40}}
\put(125,80){\line(-1,1){40}}

\put(75,50){\line(1,1){40}}
\put(65,60){\line(1,1){40}}
\put(55,70){\line(1,1){40}}

\put(115,70){\line(-1,1){40}}
\put(105,60){\line(-1,1){40}}
\put(95,50){\line(-1,1){40}}

\put(85,40){\line(0,1){80}}

\put(65,80){\circle*{3}}
\put(85,60){\circle*{3}}
\put(85,100){\circle*{3}}
\put(105,80){\circle*{3}}

\put(95,50){\circle*{3}}
\put(95,110){\circle*{3}}

\put(55,79){$\alpha$}
\put(108,79){$\negT\alpha$}
\put(66,59){{\small $\alpha\andK\negT\alpha$}}
\put(66,99){{\small $\alpha\orK\negT\alpha$}}
\put(97,47){$\alpha\andK\topT$}
\put(98,110){$\alpha\orK\topT$}

\put(83,60){\linethickness{0.5mm} \line(0,1){40}}
\put(84,101){\linethickness{0.5mm} \line(1,1){8}}
\put(84,59){\linethickness{0.5mm} \line(1,-1){8}}
\put(92,49){\linethickness{0.5mm} \line(1,1){31}}
\put(93,110){\linethickness{2.4mm} \line(1,-1){31}}

\put(100,88){{\Large M}}

\end{picture}
\end{center}
\caption{Models and cl-models.} \label{fig.1}
\end{figure}
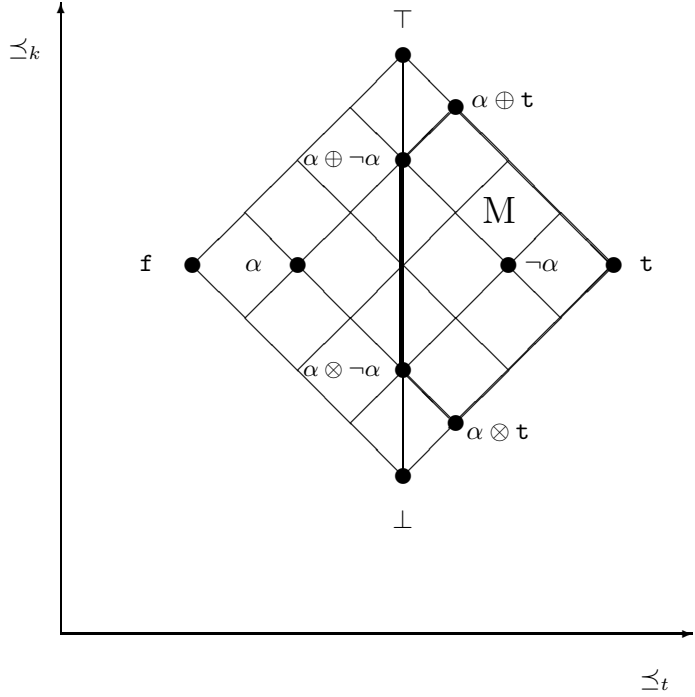
The set of models of $\lp$, $\mod{P}$, is the set of interpretations assigning to $A$ a value
in the area  ($M$-area in Figure~\ref{fig.1}) delimited by the extremal points,
$\alpha\andK\negT \alpha, \alpha\orK\negT \alpha, \alpha\orK \topT, \topT$ and
$\alpha\andK\topT$. The $\orderK$-least element $I$ of $\mod{P}$ is such that
$I(A) = \alpha\andK\topT$.

The set of cl-models of $\lp$, $\modcl{P}$, is the set of interpretations assigning
to $A$ a value on the vertical line, 
in between the extremal points $\alpha\andK\negT\alpha$ and $\alpha\orK\negT\alpha$ and are
all truth minimal. The $\orderK$-least element $I'$ of $\modcl{P}$ is such that
$I'(A) = \alpha\andK\negT \alpha$. Note that $I$ is not a cl-model of $\lp$ and,
thus, $\modcl{P} \subset \mod{P}$. \okex
\end{exmp}

\nd Clark-completion models also have an alternative characterisation. 

\begin{definition}[general reduct] \label{def.reduct}
Let $\lp$ and $I$ be a logic program and an interpretation, respectively. The \emph{general reduct}
of $\lp$ \wrt~$I$, denoted $\lp[I]$ is the program obtained from $\lpstar$ in which
each (ground) rule $A \leftarrow \varphi \in \lpstar$ is replaced with $A \leftarrow I(\varphi)$.
\okdef
\end{definition}

\nd Note that any model $J$ of $\lp[I]$ is such that for all rules $A \leftarrow \varphi \in
\lpstar$, $I(\varphi) \orderT J(A)$. But, in $\lpstar$ each ground atom appears in the head of
exactly one rule. Therefore, it is easily verified that any $\orderT$-minimal model $J$ of
$\lp[I]$ is such that $J(A) = I(\varphi)$ and there can be just one such model,
\ie~$J = \min_{\orderT}\{J' \colon J' \models \lp[I] \}$. 

We have the following theorem, which allows us to express the cl-models of a logic
program in terms of its models.

\begin{theorem} \label{theo.clmodels}
Let $\lp$ and $I$ be a logic program and an interpretation, respectively. Then
$I \modelscl \lp$ iff $I = \min_{\orderT}\{J \colon J \models \lp[I] \}$. \ok
\end{theorem}
\begin{proof}
$I \modelscl \lp$ iff for all $A \leftarrow \varphi \in \lpstar$,  $I(A) = I(\varphi)$ holds 
iff (as noted above) $I = \min_{\orderT}\{J \colon J \models \lp[I] \}$. \okproof
\end{proof}

\nd The above theorem establishes that Clark-completion models are fixed-points of
the operator $\gammap \colon {\mathcal I}(\calB) \to {\mathcal I}(\calB)$, defined as

\begin{equation} \label{op.cl}
\gammap(I) = \min_{\orderT}\{J \colon J \models \lp[I] \} \ ,
\end{equation}

\nd \ie~$I \modelscl \lp$ iff $I =\gammap(I)$.

\subsubsection{Program knowledge completions} 
Finally, given an interpretation $I$, we
introduce the
notion of program \emph{knowledge completion}, or simply, $k$-completion with $I$, denoted $\lp
\orK I$. The {\em program $k$-completion} of $\lp$ with $I$, is the program obtained by
replacing any rule of the form $A \leftarrow \varphi \in \lp$ by $A \leftarrow \varphi \orK
I(A)$.  The idea is to enforce any model  $J$ of $\lp \orK I$ to contain at least
the knowledge determined by $\lp$ and $I$. Note that  $J \models \lp \orK I$ does not imply  $J
\models \lp$. For instance, given $\lp = \{A \leftarrow A \andK \neg A\}$ and $I = J = \IbotT$,
then $\lp \orK I = \{ A \leftarrow (A \andK \neg A) \orK \botT \}$ and $J \models  \lp \orK I$,
while $J \not\models \lp$.

\subsubsection{Additional remarks}

\nd Please note that the use of the negation, $\negT$,  in literals has to be understood as
\emph{classical negation}. The expression $not~L$ (where $L$ is a literal) appearing quite
often as syntactical construct in logic programs, indicating `$L$ is not provable', is not part
of our language. This choice is intentional, as we want to stress that in this study
CWA will be considered as an additional source of (or carrier of) falsehood in an abstract
sense and will be considered as a `cumulative' information source with the classical semantics
(Kripke-Kleene semantics). In this sense, our approach is an attempt 
to relate the stable model semantics of logic programs to a standard model-theoretic account of
rules, relying on standard logical means as the notion of interpretation and knowledge ordering.
%Our approach does not depend on the presence of any specific connective, such as
%negation-as-failure, nor on any specific syntax of rules.

%In Section~\ref{sec.glp}, we will generalize our results to the case where rules have the
%general form $\varphi_{H} \leftarrow \varphi_{B}$, where both the head $\varphi_{H}$ and the
%body $\varphi_{B}$ are formulae.

%-----------------------------------
\subsection{Semantics of logic programs} \label{lpsemantics}
%-----------------------------------

\nd In logic programming, usually the semantics of a program $\lp$ is determined by selecting
a particular interpretation, or a set of interpretations, of $\lp$ in the set of models of
$\lp$.  We consider three semantics, which are probably the most popular and widely studied
semantics for logic programs with negation, namely {\em Kripke-Kleene semantics}, 
{\em well-founded semantics} and {\em stable model semantics}, in
increasing order of knowledge.
%~\cite{Fitting85,Fitting93,Fitting02,Gelfond88,vanGelder91}. 

%-----------------------------------
\subsubsection{Kripke-Kleene semantics}
%-----------------------------------

\nd Kripke-Kleene semantics~\cite{Fitting85} has a simple, intuitive and epistemic
characterization, as it corresponds to the least cl-model of a logic program under the
knowledge order $\orderK$.  Kripke-Kleene semantics is essentially a generalization of the
least model characterization of classical programs without negation over the truth space
$\{\botT,\topT\}$ (see~\cite{vanEmden76,Lloyd87}) to logic programs with classical negation
evaluated over bilattices under Clark's program completion. More formally, 

\begin{definition}[Kripke-Kleene semantics] \label{def.kk}
The  {\em Kripke-Kleene model} of a logic program $\lp$ is the $\orderK$-least cl-model of
$\lp$, \ie
 
\begin{equation} \label{kkdef} 
KK({\lp}) = \min_{\orderK}(\{I\colon I \modelscl \lp \}) \ .
\end{equation}
\okdef
\end{definition}

\nd For instance, by referring to Example~\ref{ex.models}, the value of $A$ \wrt~the Kripke-Kleene
semantics of $\lp$ is $KK({\lp})(A) = \alpha\andK\negT\alpha$.

Note that by Theorem~\ref{theo.clmodels} and by Equation~(\ref{op.cl}) we have also

\begin{equation} \label{kkdef.gamma} 
KK({\lp}) = \lfp_{\orderK}(\gammap) \ .
\end{equation}

\nd Kripke-Kleene semantics also has an alternative, and better known, fixed-point
characterization, which relies on the well-known $\phip$ immediate consequence operator. $\phip$ 
is a generalization of the Van Emden-Kowalski's immediate consequence operator 
$T_\lp$~\cite{vanEmden76,Lloyd87} to bilattices under Clark's program completion. Interesting
properties of $\phip$ are that \ii{i} $\phip$ relies on the classical evaluation of negation,
\ie~the evaluation of a negative literal $\negT A$ is given by the negation of the evaluation of $A$; and 
\ii{ii} $\phip$ is monotone with respect to the knowledge ordering and, thus, has a 
$\orderK$-least fixed-point, which coincides with the Kripke-Kleene semantics of $\lp$.
Formally,

\begin{definition}[immediate consequence operator $\phip$]\label{def.phi}
Consider a logic program $\lp$. The \emph{immediate consequence operator}  $\phip\colon
{\mathcal I}(\calB) \to {\mathcal I}(\calB)$ is defined as follows. For $I \in {\mathcal
I}(\calB)$, $\phip(I)$ is the interpretation, which for any ground atom $A$ such that $A
\leftarrow \varphi$ occurs in $\lpstar$, satisfies $\phip(I)(A) = I(\varphi)$. \okdef
\end{definition} 

\nd  It can easily be shown that

\begin{theorem}[\cite{Fitting93}]\label{thm.phipmontone}
In the space of interpretations, the operator $\phip$ is monotone 
under $\orderK$, the set of fixed-points of $\phip$ is a complete lattice under $\orderK$ and,
thus, $\phip$ has a $\orderK$-least fixed-point.  Furthermore, $I$ is a cl-model of a program
$\lp$ iff $I$ is a fixed-point of $\phip$. Therefore, the Kripke-Kleene model of $\lp$ coincides
with $\phip$'s least fixed-point under $\orderK$. \ok
\end{theorem}

\nd For instance, by referring to Example~\ref{ex.models}, the set of fixed-points 
of $\phip$ coincides with the set of interpretations assigning to $A$ a value on the vertical
line, in between the extremal points $\alpha\andK\negT\alpha$ and $\alpha\orK\negT\alpha$.

The above theorem relates the model theoretic and
epistemic characterization of the Kripke-Kleene semantics to a least fixed-point characterization. 
By relying on $\phip$ we also know how to effectively compute $KK({\lp})$ as given by the
Knaster-Tarski Theorem~\ref{thm.kt}.

Please, note that from Theorem~\ref{theo.clmodels} and Equation~(\ref{op.cl}), it follows
immediately that 

\begin{corollary} \label{cor.phip}
Let $\lp$ and $I$ be a logic program and an interpretation, respectively. Then 
 $\phip(I) = \gammap(I)$. \ok
\end{corollary}
\begin{proof}
Let $I' = \gammap(I) = \min_{\orderT}(\{J \colon J \models \lp[I]\})$. Then
we have that for any ground atom $A$,
$\gammap(I)(A) = I'(A) = I(\varphi) = \phip(I)(A)$, \ie~$\phip(I) = \gammap(I)$. \okproof
\end{proof}

\nd As a consequence, 
\emph{all definitions and properties given in this paper in terms of $\phip$
and/or cl-models may be given in terms of $\gammap$ and/or models as well.}
As $\phip$ is a well-known operator, for ease of presentation we will continue use it.

We conclude this section with the following simple lemma, which will be used later in the
paper.

\begin{lemma} \label{lem.A}
Let $\lp$ be a logic program and let $J$ and $I$ be interpretations. Then
$\phipx{\lp \orK I}(J) = \phip(J) \orK I$. In particular, $J \modelscl \lp \orK I$ iff
$J = \phip(J) \orK I$.\ok
\end{lemma}

%------------------------------------------------------------
\subsubsection{Stable model and well-founded semantics}\label{smwfs}
%------------------------------------------------------------

\nd A commonly accepted approach towards provide a stronger semantics or a semantics
that is more informative to logic programs than the Kripke-Kleene semantics, consists in relying on CWA
to complete the available knowledge. Of the various approaches to the management of negation in logic 
programming, the {\em stable model semantics} approach, introduced by Gelfond 
and Lifschitz~\cite{Gelfond88} with respect to the classical two valued truth space
$\{\botT,\topT\}$ has become one of the most widely studied and most commonly accepted proposal.
Informally, a set of ground atoms $I$ is a {\em stable model} 
of a classical logic program $\lp$ if $I = I'$, where $I'$ is computed 
according to the so-called {\em Gelfond-Lifschitz transformation}:

\begin{enumerate}
%\item if an atom is head of no rule in $\lp$ then its truth value is false ($\botT$);
\item substitute (fix) in $\lpstar$ the negative literals by their 
evaluation with respect to $I$. Let $\lp^{I}$ be the resulting {\em positive} program,
called \emph{reduct} of $\lp$ \wrt~$I$;
\item let $I'$ be the minimal Herbrand (truth-minimal) model of $\lp^{I}$.
\end{enumerate}

\nd This approach defines a whole family of models and it has been shown 
in~\cite{Przymusinski90} that the {\em minimal} one according to the 
knowledge ordering corresponds to the {\em well-founded
semantics}~\cite{vanGelder91}.  

The extension of the notions of stable model and well-founded semantics to
the context of bilattices is due to Fitting~\cite{Fitting93}.
He proposes a generalization of the Gelfond-Lifschitz 
transformation to bilattices by means of the binary immediate consequence
operator $\psip$.  Similarly to that of the 
Gelfond-Lifschitz transformation, the basic principle of $\psip$ is to separate the roles of 
positive and negative information. Informally, $\psip$ accepts two input
interpretations over a bilattice, the first is used to assign meanings to
positive literals, while the second is used to assign meanings to 
negative literals. $\psip$ is monotone  in both arguments in the 
knowledge ordering $\orderK$. But, with respect to the truth ordering $\orderT$, 
$\psip$ is monotone in the first argument, while it is antitone in the second argument (indeed, 
as the truth of a positive literal increases, the truth of its 
negation decreases). Computationally, Fitting follows the idea of the Gelfond-Lifschitz 
transformation shown above: the idea is to fix an interpretation for negative
information and to compute the $\orderT$-least model of the resulting positive
program.  To this end, Fitting~\cite{Fitting93} 
additionally introduced the $\psiprimep$  operator, which for a given interpretation $I$ of
negative literals,  computes the $\orderT$-least model, $\psiprimep(I) = \lfp_{\orderT} (\lambda
x.\psip(x,I))$. The fixed-points of $\psiprimep$ are the stable models, while the least 
fixed-point of $\psiprimep$ under $\orderK$ is the  well-founded semantics of $\lp$.  

Formally, let $I$ and $J$ be two interpretations in the bilattice 
$\tuple{{\mathcal I}(\calB), \orderT, \orderK}$. The notion of \emph{pseudo-interpretation}
$I\bigtriangleup J$ over the bilattice is defined as
follows ($I$ gives meaning to positive literals, while $J$ gives meaning to negative literals): 
for a pure ground atom $A$:
\[
\begin{array}{lcl}
(I\bigtriangleup J)(A) & = & I(A) \\
(I\bigtriangleup J)(\negT A) & = & \negT J(A) \ .
\end{array}
\]

\nd Pseudo-interpretations are extended to non-literals in the obvious way. We can now define $\psip$ as follows.

\begin{definition}[immediate consequence operator $\psip$]\label{def.psi}
The \emph{immediate consequence operator} $\psip\colon {\mathcal I}(\calB) \times {\mathcal I}(\calB) \to {\mathcal
I}(\calB)$ is defined as follows.  For $I,J \in {\mathcal I}(\calB)$,
$\psip(I,J)$ is the interpretation, which for any ground atom $A$ such that $A
\leftarrow \varphi$ occurs in $\lpstar$, satisfies $\psip(I,J)(A) = (I\bigtriangleup J)(\varphi)$. \okdef
\end{definition}

\nd
Note that $\phip$ is a special case of $\psip$, as from construction $\phip(I) = \psip(I,I)$.

The following theorem can be shown.

\begin{theorem}[\cite{Fitting93}]\label{thm.phipsi}
In the space of interpretations the operator $\psip$ is monotone in 
both arguments under $\orderK$, and under the ordering $\orderT$ it 
is monotone in its first argument and antitone in its second 
argument. \ok
\end{theorem}

\nd We are ready now to define the $\psiprimep$ operator.

\begin{definition}[stability operator $\psiprimep$]\label{def.psiprime}
The \emph{stability operator} of $\psip$ is the
single input operator $\psiprimep$ given by: $\psiprimep(I)$ is
the $\orderT$-least fixed-point of the operator
$\lambda x.\psip(x,I)$, \ie~$\psiprimep(I) = \lfp_{\orderT} (\lambda
x.\psip(x,I))$.
\okdef
\end{definition}

\nd By Theorem~\ref{thm.phipsi}, $\psiprimep$ is well defined and can be 
computed in the usual way: let $I$ be an interpretation. Consider the following sequence: for
$i \geq 0$,
\begin{eqnarray*}
v^I_0     & = & \IbotT \ ,\\
v^I_{i+1} & = & \psip(v^I_i,I) \ .
\end{eqnarray*}

\nd Then the $v^{I}_i$ sequence is monotone non-decreasing under $\orderT$ and 
converges to $\psiprimep(I)$.  In the following, with $v^I_{i}$ we will always 
indicate the $i$-th iteration of the computation of $\psiprimep(I)$.

The following theorem holds.

\begin{theorem}[\cite{Fitting93}]\label{thm.psiprime1}
The operator $\psiprimep$ is monotone in the $\orderK$ ordering,
and antitone in the $\orderT$ ordering. Furthermore, every fixed-point of $\psiprimep$ is also a
fixed-point of $\phip$, \ie~a cl-model of $\lp$. \ok
\end{theorem}

\nd Finally, following Fitting's formulation, 

\begin{definition}[stable model] \label{def.stable}
A \emph{stable model} for a logic program $\lp$ is a fixed-point of $\psiprimep$.
With $\stable{\lp}$ we indicate the set of stable models of $\lp$.  \okdef
\end{definition}
 
\nd Note that it can be seen immediately from the definition of $\psiprimep$ that

\[
\psiprimep(I) = \min_{\orderT}(\mod{P^{I}})~\footnote{As $\lp^{I}$ is positive, it has a
unique truth-minimal model.} 
\]

\nd and, thus,

\begin{equation} \label{stabledef.alt}
I \in \stable{\lp} \mbox{ iff }  I = \min_{\orderT}(\mod{{\lp^{I}}}) \ .
\end{equation}

\nd By Theorem~\ref{thm.psiprime1} and the Knaster-Tarski Theorem~\ref{thm.kt},
the set of fixed-points of $\psiprimep$, \ie~the set of stable models of $\lp$, is a complete
lattice under $\orderK$ and, thus, $\psiprimep$ has a $\orderK$-least fixed-point, which is
denoted $\lfpK$.  $\lfpK$ is known as the \emph{well-founded model}  of $\lp$ 
and, by definition, coincides with the $\orderK$-least stable model, \ie
\begin{equation}\label{lfpdef}
\lfpK = \min_{\orderK}(\{I\colon I \mbox{ stable model of } \lp \}) \ .
\end{equation}

\nd The characterization of the well-founded model in terms of least fixed-point of 
$\psiprimep$ also gives us a way to effectively compute it.

It is interesting to note, that for classical
logic programs the original definition of well-founded semantics is based on the well-known
notion of \emph{unfounded set}~\cite{vanGelder91}. The underlying principle
of the notion of unfounded sets is to identify the set of atoms that can safely be assumed
false if the current information about a logic program is given by an interpretation $I$.
Indeed, given a classical interpretation $I$ and a classical logic program $\lp$, a set of
ground atoms $X \subseteq B_\lp$ is an \emph{unfounded set} (\ie, the atoms in $X$ can be assumed
as false) for $\lp$ \wrt~$I$  iff for each atom $A \in X$, 

\begin{enumerate}
\item if  $A \leftarrow \varphi \in \lpstar$ (note that $\varphi = \varphi_1 \orT \ldots \orT \varphi_n$ and $\varphi_i= L_{i_1} \andT \ldots \andT L_{i_n}$), then $\varphi_{i}$ is false either
\wrt~$I$ or \wrt~$\neg.X$, for all $1 \leq i \leq n$.  
\end{enumerate}

\nd A well-known property of unfounded sets is that the union of two unfounded sets of $\lp$ \wrt~$I$ is an unfounded set as well and, thus, there is a unique \emph{greatest unfounded set} for $\lp$ \wrt~$I$, denoted by $U_\lp(I)$. 

%See Table~\ref{tab.1} for the greatest 
%unfounded sets \wrt~the models of the logic program in
%Example~\ref{runex}.

Now, consider the usual immediate consequence operator $T_\lp$, where for any ground atom $A$, 
\[
T_\lp(I)(A) = \topT  \mbox{ iff } \exists A\leftarrow\varphi \in \lpstar \mbox{ s.t. } I(\varphi) =
\topT,
\]
\nd and consider the well-founded operator~\cite{vanGelder91} over classical interpretations $I$
\begin{equation}\label{ref.wp}
W_\lp(I) = T_\lp(I) \cup \neg.U_\lp(I) \ .
\end{equation}
\nd $W_\lp(I)$ can be rewritten as $W_\lp(I) = T_\lp(I) \oplus \neg.U_\lp(I)$, by assuming
$\oplus = \cup, \otimes = \cap$ in the lattice $\langle 2^{B_\lp \cup \neg. B_\lp},
\subseteq\rangle$  (the partial order $\subseteq$ corresponds to the knowledge order
$\orderK$). Then, 

\begin{itemize}
\item the well-founded semantics is defined to be the $\orderK$-least fixed-point of $W_\lp$ in~\cite{vanGelder91}, and 
\item  it is shown in~\cite{Leone97} that the set of
\emph{total} stable models of $\lp$ coincides with the set of total fixed-points of $W_\lp$.
\end{itemize}

\nd In particular, this formulation reveals that the greatest unfounded set, $\neg.U_\lp(I)$, is
the additional ``false default knowledge", which is introduced by CWA into the usual
semantics of logic programs given by  $T_\lp$. However, $W_\lp$ does not allow partial
stable models to be identified. Indeed, there are fixed-points of $W_\lp(I)$ that are partial interpretations,
which are not stable models.

We  conclude the preliminary part of the paper with the following result that adds to Fitting's analysis that stable models 
are incomparable with each other with respect to the truth 
order $\orderT$.

\begin{theorem}\label{thm.psiprimesms1}
Let $I$ and $J$ be two stable models such that $I \neq J$. Then $I \not 
\orderT J$ and $J \not \orderT I$.  \ok
\end{theorem}
\begin{proof}
Assume to the contrary that either $I \orderT J$ 
or $J \orderT I$ holds. Without loss of generality, assume $I \orderT 
J$. By Theorem~\ref{thm.psiprime1}, $\psiprimep$ is antitone in the 
$\orderT$ ordering. Therefore, from $I \orderT J$ it follows that $J 
= \psiprimep(J) \orderT \psiprimep(I) = I$ holds and, thus, $I =J$, a 
contradiction to the hypothesis. \okproof
\end{proof}

%-----------------------------------
\section{Stable model semantics revisited} \label{ecopdef}
%-----------------------------------

\nd In the following, by relying on CWA as a source of falsehood for knowledge completion,  
we provide epistemic and fixed-point based, characterizations of the 
well-founded and stable model semantics over bilattices that are alternative to the one provided by 
Fitting~\cite{Fitting93}. We proceed in three steps. 

\ii{i} In the next section, we introduce the notion of {\em support}, denoted $\supportph$,
with respect to a  logic program $\lp$ and an interpretation $I$.  The support is a
generalization of the notion of greatest unfounded set (which determines the atoms that can be
assumed to be false) \wrt~$I$ from classical logic programming to bilattices. Intuitively, we
regard CWA as an additional source of information for \emph{falsehood} to be used to
complete $I$. The support $\supportph$ of $\lp$ \wrt~$I$ determines in a principled way the
amount, or \emph{degree}, of falsehood provided by CWA to the atom's truth that can be
added to current knowledge $I$ about the program $\lp$. It turns out that for classical logic
programs the support coincides with the negation of the greatest unfounded set, \ie~$\supportph
= \neg.U_\lp(I)$.

 \ii{ii} Any model $I$ of $\lp$ containing its support, \ie~such that $\supportph \orderK I$,
tells us that the additional source of falsehood provided by CWA cannot contribute
improving  our knowledge about the program $\lp$. We call such models \emph{supported models} of $\lp$; this
will be discussed in Section~\ref{s.supportmodel}. Supported models can be characterized
as fixed-points of the operator
\[
\ecopB(I)   =   \phip (I) \orK \supportph \ ,
\]
\nd which is very similar to the $W_\lp$ operator in Equation~(\ref{ref.wp}), but generalized
to bilattices. As expected, it can be shown that the \emph{$\orderK$-least supported model is
the well-founded model of $\lp$}. Unfortunately, while for classical logic programs and total
interpretations, supported models characterize total stable models (in fact, they coincides
with the fixed-points of  $W_\lp$), this is not true in the general case of interpretations over
bilattices.

Therefore,  we further refine the class of supported models, by introducing the class of \emph{models deductively closed under support k-completion}.  This class requires
supported models to satisfy some minimality condition with respect to the knowledge order
$\orderK$. Indeed, such a model $I$ has to be deductively closed according
to the Kripke-Kleene semantics of the program $k$-completed with its support, \ie

\begin{equation} \label{eq.kkstabel}
I = KK(\lp\orK \supportph)
\end{equation}
 
\nd or, equivalently

\begin{equation} \label{eq.kkstabelB}
I = \min_{\orderK}(\modcl{\lp\orK \supportph}. 
\end{equation}

 \ii{iii} We will show that any such interpretation $I$ is a stable model
of $\lp$ and vice-versa, \ie~$I \in \stable{\lp}$ iff $I = \min_{\orderK}(\modcl{\lp\orK
\supportph}$, which is quite suggestive. Note that until now, stable models (over bilattices)
have been characterized as by Equation~(\ref{stabledef.alt}). Equation~(\ref{eq.kkstabelB}) above shows thus
that stable models can be characterized as those models that contain their support and are
deductively closed under the Kripke-Kleene semantics. As such, we can identify the support
(unfounded set, in classical terms) as the added-value (in terms of knowledge), which is
brought into by the stable model semantics with respect to the standard Kripke-Kleene semantics
of $\lp$. 

%Indeed, an interpretation $I$
%is a stable model of $\lp$ iff for 
%any rule $A \leftarrow \varphi \in \lpstar$ we have that
%$I(A) = I(\varphi) \orK \supportph(A)$, meaning 
%that the information about $A$ is given by
%the information provided by $\varphi$ joined 
%together with the maximal amount of falsehood
%provided by the CWA to $A$. 

Finally, stable models can thus be defined in terms of fixed-points of the operator 
$KK(\lp \orK \supportphIx{\cdot})$, which relies on a, though intuitive, program transformation
$\lp\orK \supportphIx{\cdot}$.  We further introduce a new operator $\phiprimep$, which we show to have the property that $\phiprimep(I) = KK(\lp \orK \supportph)$. This operator clearly shows that
a model is a stable model iff it contains exactly the knowledge obtained by activating
the rules over its support, without any other extra knowledge. An important property of 
$\phiprimep$ is that it does depend on $\phip$ only.  This may be important in the classical
logic programming case where $\lp\orK\supportphIx{\cdot}$ is not easy to define (as $\orK$
does not belong to the language of classical logic programs). As a consequence,  no program transformation is required, which completes our
analysis.  

We will rely on the following running example to illustrate the concepts that will be introduced in the next sections.

\begin{exmp}[running example] \label{runex}
Consider the following logic program $\lp$ with the following rules.
\[
\begin{array}{l}
p \leftarrow p \\
q \leftarrow \negT r \\
r \leftarrow \negT q \andT \negT p \\
\end{array}
\]

\nd In Table~\ref{tab.1} we report the cl-models $I_{i}$, the Kripke-Kleene, the well-founded
and the stable models of $\lp$, marked by bullets. Note that according to
Theorem~\ref{thm.psiprimesms1}, stable models cannot be compared with each other under
$\orderT$, while under the knowledge order, $I_{3}$ is the least informative model (\ie~the
well-founded model), while $I_{6}$ is the most informative one ($I_{4}$ and $I_{5}$ are
incomparable under $\orderK$).
\begin{table}
\caption{Models, Kripke-Kleene, well-founded and stable models of $\lp$.} \label{tab.1}
\[
{\footnotesize 
%\begin{array}{|c|ccc|c|c|c|} \hline
\begin{array}{ccccccc} \hline\hline
I_{i} \modelscl \lp & p & q & r & KK(\lp) & WF(\lp) & \mbox{stable models} \\ \hline
I_{1} & \botK & \botK & \botK & \bullet &         &        \\ %\hline
I_{2} & \botK & \topT & \botT &         &         &        \\ %\hline
I_{3} & \botT & \botK & \botK &         & \bullet & \bullet \\ %\hline
I_{4} & \botT & \botT & \topT &         &         & \bullet \\ %\hline
I_{5} & \botT & \topT & \botT &         &         & \bullet \\ %\hline
I_{6} & \botT & \topK & \topK &         &         & \bullet \\ %\hline
I_{7} & \topT & \topT & \botT &         &         &         \\ %\hline
I_{8} & \topK & \topT & \botT &         &         &         \\ %\hline
I_{9} & \topK & \topK & \topK &         &         &         \\ \hline\hline
\end{array}
}
\]
\end{table}
\okex
\end{exmp}

%--------
%\subsection{minimal supported semantics (model theory approach)} \label{s.model}
\subsection{Support} \label{s.support}

\nd The main notion we introduce here is that of {\em support} of a logic program $\lp$ with
respect to a given interpretation $I$. If $I$ represents what we already know about an intended
model of $\lp$, the support represents the $\orderK$-greatest amount/degree of falsehood
provided by CWA that can be joined to $I$ in order to complete $I$. Falsehood is always represented in
terms of an interpretation, which we call a \emph{safe} interpretation. The main principle
underlying safe interpretations can be explained as follows. For ease, let us consider ${\cal
FOUR}$. Consider an interpretation $I$, which is our current knowledge about $\lp$.   Let us
assume that the interpretation $J$,  with $J \orderK \IbotT$,  indicates which atoms may be
assumed as $\botT$. For any ground atom $A$, $J(A)$ is the default `false' information provided
by $J$ to the atom $A$. The completion of $I$ with $J$ is the interpretation $I \orK J$.  In
order to accept this completion, we have to ensure that  the assumed false knowledge about $A$,
$J(A)$, is entailed by $\lp$ \wrt~the completed interpretation $I \orK J$, \ie~for $A
\leftarrow \varphi \in \lpstar$, $J(A) \orderK (I\orK J)(\varphi)$ should hold. 
That is, after completing the current knowledge $I$ about $\lp$ with the `falsehood'
assumption $J$, the inferred information about $A$, $(I\orK J)(\varphi)$, should increase. Formally,

\begin{definition}[safe interpretation] \label{def.safe}
Let $\lp$ and $I$ be a logic program and an interpretation, respectively. An 
interpretation $J$ is \emph{safe} \wrt~$\lp$ and $I$ iff:
\begin{enumerate}
\item $J \orderK \IbotT$;
\item $J \orderK \phip(I \orK J)$.  \okdef
\end{enumerate}
\end{definition}

\nd As anticipated, safe interpretations have an interesting reading once we restrict our attention to the
classical framework of logic programming: indeed, the concept of safe interpretation reduces to
that of unfounded set. 

\begin{theorem} \label{thm.ufs}
Let $\lp$ and $I$ be a  classical logic program and a classical interpretation, respectively.
Let $X$ be a subset of $B_{\lp}$. Then $X$ is an unfounded set of $\lp$ \wrt~ $I$ iff $\neg.X
\orderK \phip(I \oplus \neg.X)$~\footnote{Note that this condition can be rewritten as $\neg.X
\subseteq \phip(I \cup \neg.X)$.}, \ie~$\neg. X$ is safe  \wrt~$\lp$ and $I$.\ok
\end{theorem}
\begin{proof}
Assume $\negT A \in \neg.X$ ( \ie~$\neg.X(\negT A) = \topT$) and, thus, $A \in X$ (\ie~$X(A) = \botT$). Therefore, by definition of unfounded sets, if  $A \leftarrow \varphi \in \lpstar$, where $\varphi = \varphi_1 \orT \ldots \orT \varphi_n$ and $\varphi_i= L_{i_1} \andT \ldots \andT L_{i_n}$, then either $I(\varphi_{i}) = \botT$ or $\neg.X(\varphi_{i}) = \botT$. Therefore, $(I \cup \neg.X)(\varphi) = \botT$, \ie~$(I \oplus \neg.X)(\varphi) = \botT$. But then, by definition of $\phip$, we have that $\phip(I \oplus \neg.X)(A) = \botT$, \ie~$\phip(I \oplus \neg.X)(\negT A) = \topT$. Therefore, $\neg.X \orderK \phip(I \oplus \neg.X)$.
The other direction can be shown similarly. \okproof
\end{proof}

\nd The following example illustrates the concept.

\begin{exmp}[running example cont.] \label{runex.1}
Let us consider $I_{2}$. $I_{2}$ dictates that $p$ is unknown, $q$ is true and that $r$ 
is false. Consider the interpretations $J_{i}$ defined as follows:
\[
\begin{array}{cccc} \hline\hline
J_{i} & p & q & r \\ \hline
J_{1} & \botK & \botK & \botK \\ %\hline
J_{2} & \botT & \botK & \botK \\ %\hline
J_{3} & \botK & \botK & \botT \\ %\hline
J_{4} & \botT & \botK & \botT \\  \hline\hline
\end{array}
\]

\nd It is easy to verify that $J_{i} \orderK \IbotT$ and $J_{i} \orderK \phip(I_{2}
\orK J_{i})$. Therefore, all the $J_{i}$s are safe. The $\orderK$-least safe interpretation
is $J_{1}$, while the $\orderK$-greatest safe interpretation is $J_{4} = J_{1} \orK J_2 \orK
J_3$. $J_{4}$ dictates that under $I_{2}$, we can `safely' assume that both $p$ and $r$ are
false. Note that if we join $J_{4}$ to $I_{2}$ we obtain the stable model $I_{5}$, where $I_{2}
\orderK I_{5}$. Thus, $J_{4}$ improves the knowledge expressed by
$I_{2}$.

It might be asked why we do not consider $q$ false as well. In fact, if we consider $p,q$
and  $r$ false, after joining to $I$ and applying $\phip$,  $q$ becomes true, which
is \emph{knowledge-incompatible} with $q$'s previous knowledge status ($q$ is false). So,
$q$'s falsehood is not preserved. \okex
\end{exmp}

\nd We also consider another example on a more general bilattice allowing the management of uncertainty.

\begin{exmp} \label{exmy}
Let us consider the lattice $\tuple{L,\preceq}$, where $L$ is the unit interval $[0,1]$ and
$\preceq$ is the natural linear order $\leq$. The negation operator on $L$ considered is
defined as $\negT x = 1- x$. We further build the bilattice of intervals
\mbox{$\tuple{[0,1] \times [0,1],\orderT,\orderK}$} in the standard way. An interval
$\tuple{x,y}$ may be understood as an approximation of the certainty of an atom.

Let us note that for $x,x',y,y' \in L$,
\begin{itemize}
\item $\tuple{x,y} \andT \tuple{x',y'} = \tuple{\min(x,x'),\min(y,y')}$;
\item $\tuple{x,y} \orT \tuple{x',y'} = \tuple{\max(x,x'),\max(y,y')}$;
\item $\tuple{x,y} \andK \tuple{x',y'} = \tuple{\min(x,x'),\max(y,y')}$; 
\item $\tuple{x,y} \orK \tuple{x',y'} = \tuple{\max(x,x'),\min(y,y')}$; and
\item $\neg \tuple{x,y} = \tuple{1-y,1-x}$.
\end{itemize}

Consider the logic program $\lp$ with rules
\[
\begin{array}{l}
A \leftarrow A \andT C \\
B \leftarrow B \orT \neg C \\
C \leftarrow C \orT D \\
D \leftarrow [0.7,0.7]
\end{array}
\]

\nd The fourth rule asserts that the truth value of $D$ is exactly 0.7.
Then using the third rule, we will infer that the value of $C$ is given
by the disjunction of 0.7 and the value of $C$ itself which is initially unknown, \ie~between
0 and 1, thus our knowledge about $C$ is that its value is at least 0.7,
\ie~[0.7;1]. Activating the second rule with that knowledge, then the value 
of $B$ is given by the disjunction of the value of $\neg C$, 
that is at most 0.3, \ie~[0;0.3], and the value of $B$ itself that is unknown, thus $B$ remains unknown.
Similarly, the first rule does not provide any knowledge about the value of $A$. That knowledge corresponds to the Kripke-Kleene model $I$ of $\lp$, obtained by iterating $\phip$
starting with $\IbotK$ : $I(A) = [0;1]$, $I(B) = [0;1]$, $I(C) = [0.7;1]$ and $I(D) = [0.7;0.7]$.  

Relying on CWA, we should be able to provide a more precise characterization
of $A$, $B$ and $C$. It can be verified that it may be safely assumed that $A$ is false ($[0;0]$)
and that the truth of $B$ and $C$ is at most 0.3 and 0.7, respectively, which combined with $I$ determines a more precise interpretation where $A$ is false, $B$ is at most $0.3$, $C$ is $0.7$ and $D$ is $0.7$, respectively, as highlighted in the following table. Consider interpretations $I,J_1, J_2, J'$.
\[
\begin{array}{ccccc} \hline\hline
      & A & B & C  & D  \\  \hline
  I    & [0;1] & [0;1] & [0.7;1] & [0.7;0.7]  \\ %\hline
  J_1    & [0;0]& [0;1] & [0;0.8] & [0;0.7]   \\ %\hline
  J_2    & [0;1]& [0;0.3] & [0;0.7] & [0;1]  \\ %\hline
  J'    & [0;0]& [0;0.3] & [0;0.7] & [0;0.7] \\ %\hline
  I \orK J'    & [0;0]& [0.0;0.3] & [0.7;0.7] & [0.7;0.7] \\ \hline\hline
\end{array}
\]

\nd Both $J_1$ and $J_2$ are safe \wrt~$\lp$ and $I$.  It is easy to see that 
  $J' = J_1 \orK  J_2$ is the $\orderK$-greatest
safe interpretation. Interestingly, note how
$J'$ provides to $I$  some additional information  on the values
of $A$, $B$ and $C$, respectively.
\okex
\end{exmp}

\nd Of all possible safe interpretations \wrt~$\lp$ and $I$, we are interested in the
maximal one under $\orderK$, which is unique. The $\orderK$-greatest safe interpretation 
will be called the support provided by CWA to $\lp$ \wrt~$I$.

\begin{definition}[support] \label{def.support}
Let $\lp$ and $I$ be a logic program and an interpretation, respectively. The \emph{support provided by CWA} to $\lp$ \wrt~$I$, or simply \emph{support}
of $\lp$ \wrt~$I$, denoted $\supportph$, is the $\orderK$-greatest safe
interpretation \wrt~$\lp$ and $I$, and is given by
\[
\supportph = \bigorK \{ J \colon J \mbox{ is safe \wrt\ }
\lp \mbox{ and } I  \} \ .  
\]
\okdef
\end{definition}

\nd It is easy to show that support is a well-defined concept. Consider $X = \{ J \colon J$ is safe \wrt\ $\lp$ and  $I  \}$. As the bilattice is a complete lattice under $\orderK$,
$lub_{\orderK}(X) = \orK_{J \in X} J$ and, thus, by definition $\supportp(I) =  lub_{\orderK}(X)$.
Now consider $J \in X$.  Therefore $J \orderK \supportp(I)$. But $J$ is safe, so
$J \orderK \IbotT$ and $J \orderK \phip(I \orK J) \orderK \phip(I \orK \supportp(I))$ (by $\orderK$-monotonicity of $\phip$). As a consequence, both $\IbotT$ and $\phip(I \orK \supportp(I))$ are upper bounds of $X$. But $\supportp(I)$ is the least upper bound of $X$ and, thus,
$\supportp(I) \orderK \IbotT$ and $\supportp(I) \orderK \phip(I \orK \supportp(I))$ follows. That is, $\supportp(I)$ is safe and the $\orderK$-greatest safe interpretation \wrt~$\lp$ and~$I$. 

%Given two safe
%interpretations $J$ and $J'$, then $J \orK J' \orderK \IbotT$ and, from the monotonicity of
%$\phip$ under $\orderK$, $J \orK J'\orderK \phip(I \orK J  \orK J')$ and, thus, $J \orK J'$ is
%safe. Therefore, $\bigorK \{ J \colon J \mbox{ is safe \wrt\ } \lp \mbox{ and } I  \} \ $ is the
%$\orderK$-greatest safe interpretation \wrt~$\lp$ and~$I$. 

It follows immediately from Theorem~\ref{thm.ufs} that, in the classical setting, the
notion of greatest unfounded set is captured by the notion of support, \ie~the support tells us
which atoms may be safely assumed to be false, given a classical interpretation $I$ and a
classical logic program $\lp$.  Therefore, the notion of support extends the notion of greatest unfounded 
sets from the classical setting to bilattices. 

\begin{corollary} \label{cor.ufs} Let $\lp$ and $I$ be a  classical logic program and a
classical interpretation, respectively. Then $\supportph = \neg.U_\lp(I)$. \ok 
\end{corollary}

\begin{exmp}[running example cont.] \label{runex.2}
Table~\ref{tab.2} extends Table~\ref{tab.1} also by including the supports $\supportphIx{I_{i}}$. Note that, according to Corollary~\ref{cor.ufs}, $\supportphIx{I_{i}} = \neg.U_\lp(I_{i})$.
\begin{table}
\caption{Running example cont.: support of $\lp$ \wrt~$I_{i}$.} \label{tab.2}
\[
%{\small
{\footnotesize 
%\begin{array}{|c||ccc||ccc|c||c|c|c|} \hline
\begin{array}{ccccccccccc} \hline\hline
&  \multicolumn{3}{c}{I_{i}} & \multicolumn{3}{c}{\supportphIx{I_{i}}} &   & &  & \mbox{stable} \\ 
I_{i} \modelscl \lp & p & q & r & p & q & r & U_\lp(I_{i}) & KK(\lp) & WF(\lp) & \mbox{models} \\ \hline
I_{1} & \botK & \botK & \botK & \botT & \botK & \botK &  \{p\} & \bullet &         &        \\ %\hline
I_{2} & \botK & \topT & \botT & \botT & \botK & \botT &  \{p,r\} &       &         &        \\ %\hline
I_{3} & \botT & \botK & \botK & \botT & \botK & \botK &  \{p\}  &      & \bullet & \bullet \\ %\hline
I_{4} & \botT & \botT & \topT & \botT & \botT & \botK &  \{p,q\} &       &         & \bullet \\ %\hline
I_{5} & \botT & \topT & \botT & \botT & \botK & \botT &  \{p,r\} &       &         & \bullet \\ %\hline
I_{6} & \botT & \topK & \topK & \botT & \botT & \botT &   \{p,q,r\}&       &         & \bullet \\ %\hline
I_{7} & \topT & \topT & \botT & \botT & \botK & \botT &   \{p,r\}&       &         &         \\ %\hline
I_{8} & \topK & \topT & \botT & \botT & \botK & \botT &   \{p,r\}&       &         &         \\ %\hline
I_{9} & \topK & \topK & \topK & \botT & \botT & \botT &   \{p,q,r\}&      &         &         \\ \hline\hline
\end{array}
}
\]
\end{table}
\okex
\end{exmp}

\nd 
Having defined the support model-theoretically, we next show how the support can
effectively be computed as the iterated fixed-point of a function, $\supportfunph$, that depends
on $\phip$ only. Intuitively, the iterated computation weakens $\IbotT$, \ie~CWA,
until we arrive to the $\orderK$-greatest safe interpretation, \ie~the support.

\begin{definition}[support function] \label{def.supportfun}
Let $\lp$ and $I$ be a logic program and an interpretation, respectively.
The \emph{support function}, denoted $\supportfunph$, \wrt~$P$ and $I$ is the
function mapping interpretations into interpretations defined as follows: for
any interpretation $J$,
\begin{eqnarray*}
\supportfunph(J) &  = & \IbotT \andK \phip(I \orK J) \ .
\end{eqnarray*}
\okdef
\end{definition}

\nd It is easy to verify that $\supportfunph$ is monotone \wrt~$\orderK$.
The following theorem determines how to compute the support.

\begin{theorem}\label{thm.supportcomp}
Let $\lp$ and $I$ be a logic program and an interpretation, 
respectively.  Consider the iterated sequence of
interpretations $\hypohI_{i}$ defined as follows: for any $i \geq 0$,
\[
\begin{array}{lcl}
\hypohI_{0} & = & \IbotT \ , \\ \\
\hypohI_{i+1} & = & \supportfunph(\hypohI_{i}) \ .
\end{array}
\]

\nd The sequence $\hypohI_{i}$ is 

\begin{enumerate}
\item monotone non-increasing under $\orderK$ and, thus, reaches a fixed-point
$\hypohI_{\lambda}$, for a limit ordinal $\lambda$; and

\item is monotone non-decreasing under $\orderT$.
\end{enumerate}
Furthermore,  $\supportph = \hypohI_{\lambda}$ holds. \ok
\end{theorem}
\begin{proof}
The proof is by induction. Concerning Point 1., $\hypohI_{1} \orderK \hypohI_{0}$; for all successor ordinal i, as $\supportfunph$ is monotone under $\orderK$, if $\hypohIx{I}_{i+1} \orderK \hypohIx{I}_{i}$
then $\hypohIx{I}_{i+2} \orderK \hypohIx{I}_{i+1}$; and for all limit ordinal $\lambda$,
if $i < \lambda$ then $\hypohIx{I}_{\lambda} =  \bigotimes_{i < \lambda} \hypohIx{I}_{i} 
\orderK \hypohIx{I}_{i}$. Thus the sequence is monotone 
non-increasing under $\orderK$. Therefore, the sequence has a fixed-point 
at the limit, say $\hypohI_{\lambda}$. 

Concerning Point 2., $\hypohI_{0} \orderT \hypohI_{1}$; for all successor ordinal i, from $\hypohIx{I}_{i+1} \orderK \hypohIx{I}_{i}$, by Lemma~\ref{prop.p9}, we have $\hypohIx{I}_{i} = \hypohIx{I}_{i} \andK \IbotT \orderT \hypohIx{I}_{i+1}$; and similarly, for all limit ordinal $\lambda$, if
$i < \lambda$, we have $\hypohIx{I}_{\lambda} \orderK \hypohIx{I}_{i}$, and by Lemma~\ref{prop.p9}, $\hypohIx{I}_{i} = \hypohIx{I}_{i} \andK \IbotT \orderT \hypohIx{I}_{\lambda}$.

Let us show that 
$\hypohI_{\lambda}$ is safe and $\orderK$-greatest.
$\hypohI_{\lambda} = \supportfunph(\hypohI_{\lambda}) = \IbotT \andK 
\phip(I \orK \hypohI_{\lambda})$. Therefore,  $\hypohI_{\lambda} 
\orderK \IbotT$ and $\hypohI_{\lambda} \orderK \phip(I \orK 
\hypohI_{\lambda})$, so $\hypohI_{\lambda}$ is safe \wrt~$\lp$ and $I$.

Consider any $X$ safe \wrt~$\lp$ and $I$. We show by induction on $i$ that
$X \orderK \hypohI_{i}$ and, thus, at the limit $X \orderK 
\hypohI_{\lambda}$, so $\hypohI_{\lambda}$ is $\orderK$-greatest.

\ii{i} Case $i=0$. By definition, $X \orderK \IbotT = \hypohI_{0}$.

\ii{ii} Induction step: suppose $X \orderK \hypohI_{i}$. Since $X$ is 
safe, we have $X \orderK X \andK X \orderK \IbotT \andK \phip(I \orK 
X)$. By induction, using the monotonicity of $\supportfunph$ \wrt~$\orderK$, 
$X \orderK \IbotT \andK \phip(I \orK \hypohI_{i}) = \hypohI_{i+1}$.

\ii{iii} Transfinite induction: given an ordinal limit $\lambda$, 
suppose $X \orderK \hypohI_{i}$ holds for all $i < \lambda$.
Using the fact that the space of interpretations $\tuple{{\mathcal
I}(\calB), \orderT, \orderK}$ is an infinitary interlaced bilattice,
we have $X \orderK \bigotimes_{i < \lambda} \hypohIx{I}_{i} =  \hypohI_{\lambda}$,
which concludes the proof. \okproof
\end{proof}

\nd In the following, with $\hypohIx{I}_{i}$ we indicate
the $i$-th iteration of the computation of the support of $\lp$ \wrt~$I$, according to 
Theorem~\ref{thm.supportcomp}.

Note that by construction
\begin{equation} \label{eq.supporteq}
\supportph = \IbotT \andK \phip(I \orK \supportph) \ ,
\end{equation}

\nd which establishes also that the support is deductively closed in terms of falsehood. In fact, even 
if we add all that we know about the atom's falsehood to the current interpretation $I$, we know no more about the atom's falsehood than we knew before. 

Interestingly, for a classical logic program $\lp$ and a classical interpretation $I$, by
Corollary~\ref{cor.ufs}, the above method gives us a simple top-down method to compute the
negation of the greatest unfounded set, $\neg. U_\lp(I)$, as the limit of the sequence:

\[
\begin{array}{lcl}
\hypohI_{0}  & = &  \neg.B_\lp \ , \\
\hypohI_{i+1} & = & \neg.B_\lp \cap \phip(I \cup \hypohI_{i}) \ .
\end{array}
\]

\nd The support $\supportph$ can be seen as an operator over the space of interpretations. The following 
theorem asserts that the support is monotone \wrt~$\orderK$.

\begin{theorem}\label{thm.supportmonotone}
Let $\lp$ be a logic program. The support operator $\supportp$ 
is monotone \wrt~$\orderK$. \ok
\end{theorem}
\begin{proof}
Consider two interpretations $I$ and $J$, where $I \orderK 
J$. Consider the two sequences
$\hypohIx{I}_{i}$ and $\hypohIx{J}_{i}$.
We show by induction on $i$ that $\hypohIx{I}_{i} \orderK 
\hypohIx{J}_{i}$ and, thus, at the limit $\supportphIx{I} \orderK 
\supportphIx{J}$.

\ii{i} Case $i=0$. By definition, $\hypohIx{I}_{0} = \IbotT \orderK 
\IbotT = \hypohIx{J}_{0}$.

\ii{ii} Induction step: suppose $\hypohIx{I}_{i} \orderK 
\hypohIx{J}_{i}$. By monotonicity under $\orderK$ of $\phip$  and the 
induction hypothesis,
$\hypohIx{I}_{i+1} = \IbotT \andK \phip(I \orK \hypohIx{I}_{i}) 
\orderK \IbotT \andK \phip(J \orK \hypohIx{J}_{i}) = 
\hypohIx{J}_{i+1}$, which concludes. \okproof
\end{proof}

\nd The next corollary follows directly from Lemma~\ref{prop.p9}.

\begin{corollary}\label{cor.supportmonotone}
Let $\lp$ be a logic program and consider two interpretations $I$ and 
$J$ such that $I \orderK J$. Then  $\supportp(J) \orderT \supportp(I)$. \ok
\end{corollary}

%-------------------------------
%\subsection{Supported models} \label{s.supportmodel}
\subsection{Models based on the support} \label{s.supportmodel}

\nd Of all possible models of a program $\lp$, we are especially 
interested in those models $I$ that already integrate their own support, \ie~that could not
be completed by CWA.

\begin{definition}[supported model]\label{def.corrsupmod}
Consider a logic program $\lp$. An interpretation $I$ is a 
\emph{supported model} of $\lp$ iff $I \modelscl \lp$ and $\supportph \orderK I$. \okdef
\end{definition}

\nd If we consider the definition of support in the classical setting, then supported models
are classical models of classical logic programs such that $\neg.U_{\lp}(I) \subseteq I$,
\ie~the false atoms provided by the greatest unfounded set are already false in the interpretation $I$.
Therefore, CWA does not further contribute improving $I$'s knowledge about the program
$\lp$. 

\begin{exmp}[running example cont.] \label{runex.3}
Table~\ref{tab.3} extends Table~\ref{tab.2} by also including supported models. Note that while both $I_{8}$ and $I_{9}$ are models of $\lp$ including their support,
they are not stable models. Note also that $\supportphIx{I_{8}} = \supportphIx{I_{5}}$ and
 $\supportphIx{I_{9}} = \supportphIx{I_{6}}$. That is, $I_{8}$ and $I_{9}$, which are not
stable models, have the same support of some stable model.
\begin{table}
\caption{Running example cont.: supported models of $\lp$.} \label{tab.3}
\[
%{\small
{\footnotesize 
%\begin{array}{|c||ccc||ccc|c||c|c|c|c|} \hline
\begin{array}{cccccccccccc} \hline\hline
&  \multicolumn{3}{c}{I_{i}} & \multicolumn{3}{c}{\supportphIx{I_{i}}} & & &  &\mbox{stable} & \mbox{supported}\\ 
I_{i} \modelscl \lp & p & q & r & p & q & r & U_\lp(I_{i}) & KK(\lp) & WF(\lp) & \mbox{models} & \mbox{models}\\\hline
I_{1} & \botK & \botK & \botK & \botT & \botK & \botK & \{p\} & \bullet &         &  &      \\ %\hline
I_{2} & \botK & \topT & \botT & \botT & \botK & \botT & \{p,r\} &         &         &   &     \\ %\hline
I_{3} & \botT & \botK & \botK & \botT & \botK & \botK & \{p\}  &         & \bullet&\bullet & \bullet\\ %\hline
I_{4} & \botT & \botT & \topT & \botT & \botT & \botK &  \{p,q\} &         &         &\bullet & \bullet \\ %\hline
I_{5} & \botT & \topT & \botT & \botT & \botK & \botT &  \{p,r\} &         &         &\bullet &\bullet\\  %\hline
I_{6} & \botT & \topK & \topK & \botT & \botT & \botT &   \{p,q,r\}&         &         & \bullet& \bullet \\ %\hline
I_{7} & \topT & \topT & \botT & \botT & \botK & \botT &   \{p,r\}&         &         &         &\\ %\hline
I_{8} & \topK & \topT & \botT & \botT & \botK & \botT &   \{p,r\}&          &         &    & \bullet     \\ %\hline
I_{9} & \topK & \topK & \topK & \botT & \botT & \botT &   \{p,q,r\}&         &         &    & \bullet     \\ \hline\hline
\end{array}
}
\]
\end{table}
\okex
\end{exmp}

\nd Supported models have interesting properties, as stated below.

\begin{theorem}\label{thm.model2}
Let $\lp$ and $I$ be a logic program and an interpretation, respectively. The following
statements are equivalent:
\begin{enumerate}
\item $I$ is a supported model of $\lp$;
\item $I = \phip(I) \orK \supportph$;
\item $I \modelscl \lp \orK \supportph$;
\item $I =\phip(I \orK \supportph)$. \ok
\end{enumerate}
\end{theorem}
\begin{proof}
Assume Point~1.~holds, \ie~$I \modelscl \lp$ and $\supportph \orderK I$. Then, $I = \phip(I) = 
\phip(I) \orK
\supportph$, so Point~2.~holds.

Assume Point~2.~holds. Then, by Lemma~\ref{lem.A}, 
$I = \phip(I) \orK \supportph = \phipx{\lp \orK \supportph}(I)$, \ie~$I \modelscl
\lp  \orK \supportph$, so Point~3.~holds. 

Assume Point~3.~holds. So, $\supportph \orderK I$ and from the safeness of
$\supportp(I)$, it follows that $\supportp(I) \orderK \phip(I\orK \supportp(I)) =
\phip(I)$ and, thus, $I = \phipx{\lp \orK \supportph}(I) =  \phip(I) \orK \supportph =
\phip(I)$. Therefore, $\phip(I \orK \supportph) = \phip(I) = I$, so
Point~4. holds.

Finally, assume Point~4.~holds. From the safeness of
$\supportp(I)$, it follows that $\supportp(I) \orderK \phip(I\orK \supportp(I)) = I$.
Therefore, $I = \phip(I\orK \supportp(I)) = \phip(I)$ and, thus $I$ is a supported model of
$\lp$. So, Point~1.~holds, which concludes the proof.\okproof
\end{proof}

\nd The above theorem states the same concept in different ways: supported models
contain the amount of knowledge expressed by the program and their support.

From a fixed-point characterization point of view,  from Theorem~\ref{thm.model2} it
follows that the set of supported models can be identified by the fixed-points of the
$\orderK$-monotone  operators $\ecop$ and $\ecopB$ defined
by
\begin{eqnarray} \label{ecopd} 
\ecop(I) &  = & \phip (I\orK \supportp(I)) \ , \\
\ecopB(I) & = & \phip (I) \orK \supportp(I) \ . \label{ecopdd} 
\end{eqnarray}

\nd It follows immediately  that

\begin{theorem} \label{thm.ecop}
Let $\lp$ be a logic program. Then $\ecopB$ ($\ecop$) is monotone under $\orderK$. Furthermore, an
interpretation $I$ is a supported model iff $I = \ecopB(I)$ ($I = \ecop(I)$) and, thus, relying on
the Knaster-Tarski fixed-point theorem (Theorem~\ref{thm.kt}), the set of supported
models is a complete lattice under $\orderK$. \ok
\end{theorem}

\nd Note that $\ecop$ has been defined first
in~\cite{Loyer03a} without recognizing that it characterizes supported models. However, it has been
shown in~\cite{Loyer03a} that the least fixed-point under $\orderK$ 
coincides with the well-founded semantics, \ie~in our context, 
\emph{the $\orderK$-least supported model of $\lp$ is the well-founded semantics of
$\lp$}. 

\begin{theorem}[\cite{Loyer03a}]\label{thm.supmod.char}
Consider a logic program $\lp$. Then $\lfpK = \lfp_{\orderK} (\ecop)$ ($\lfpK = \lfp_{\orderK} (\ecopB)$) and stable models
are fixed-points of $\ecop$ ($\ecopB$).\ok
\end{theorem}

\begin{exmp}[running example cont.] \label{runex.5} Consider Table~\ref{tab.3}. Note that 
stable models are supported models, \ie~fixed-points of $\ecopB$ ($\ecop$), and that the
$\orderK$-least supported model coincides with the well-founded model. Additionally, $I_{8}$
and $I_{9}$ are fixed-points of $\ecopB$ ($\ecop$) and not stable models. Thus, stable models are a proper
subset of supported models.\okex
\end{exmp}

\nd Note that the above theorem is not surprising considering that the $\ecopB$
operator is quite similar to the $W_\lp$ operator defined in Equation~(\ref{ref.wp}) for
classical logic programs and interpretations. The above theorem essentially extends the
relationship to general logic programs interpreted over bilattices. But,
while for classical logical programs and total interpretations, $\ecopB(I)$
characterizes stable total models (as,  $\ecopB = W_\lp$), this is not true in the general case
of interpretations over bilattices (\eg, see Table~\ref{tab.3}).

%-------------------------------
%\subsection{Stable supported models} \label{s.supportcompletemodel}

As highlighted in Examples~\ref{runex.3} and \ref{runex.5}, supported models are
not specific enough to completely identify stable models: we must further refine the notion
of supported models. Example~\ref{runex.3} gives us a hint. For instance, consider the
supported model $I_{8}$. As already noted, the support (in classical terms, the greatest
unfounded set) of $I_{8}$ coincides with that of $I_{5}$, but for this support,
\ie~$\supportphIx{I_{5}}$, $I_{5}$ is the $\orderK$-least informative cl-model, \ie~$I_{5}
\orderK I_{8}$. Similarly, for support $\supportphIx{I_{6}}$, $I_{6}$ is the $\orderK$-least
informative cl-model, \ie~$I_{6} \orderK I_{9}$. It appears clearly that some supported models
contain knowledge that cannot be inferred from the program or from  CWA.  
This may suggest partitioning supported models into sets of cl-models with a given support and then taking the least informative one to avoid that the supported models contain unexpected extra knowledge. 

Formally, for a given interpretation $I$, we will consider the class of all models of $\lp
\orK \supportph$, \ie~interpretations which contain the knowledge entailed by $\lp$ and the support $\supportph$,
and then take the $\orderK$-least model. If this $\orderK$-least model is $I$ itself then $I$
is a  supported model of $\lp$ deductively closed under support k-completion.

\begin{definition}[model deductively closed under support k-completion]\label{def.clcommod}
Let $\lp$ and $I$ be a logic program and an interpretation, respectively. Then $I$ is a
\emph{model deductively closed under support k-completion} of $P$ iff $I = \min_{\orderK}(\supportClass{\supportph})$. \okdef
\end{definition}

\nd Note that by Lemma~\ref{lem.A}, 

\begin{equation} \label{eq.classequiv}
\supportClass{\supportph} = \{J\colon J = \phip(J) \orK \supportph \} \ .
\end{equation}

\nd Therefore, if $I$ is a model deductively closed under support k-completion 
then $I = \phip(I) \orK \supportph$,
\ie~$I\modelscl \lp \orK \supportph$. Therefore, by Theorem~\ref{thm.model2}, any model deductively closed under support k-completion is also a supported model, \ie~$I\modelscl \lp$ and $\supportph \orderK I$.

Interestingly, models deductively closed under support k-completion have also a 
different, equivalent and quite suggestive
characterization. In fact, from the definition it follows immediately that

\begin{eqnarray} \label{eqA}
\min_{\orderK}(\supportClass{\supportph}) & = &  KK(\lp \orK \supportph) \ . \nonumber
\end{eqnarray}

\nd It then follows that

\begin{theorem} \label{thm.equivalence}
Let $\lp$ and $I$ be a logic program and an interpretation, respectively. Then $I$ is a 
model deductively closed under support k-completion
of $\lp$ iff $I = KK(\lp\orK\supportph)$.\ok
\end{theorem}

\nd That is, given an interpretation $I$ and logic program $\lp$, 
among all cl-models of $\lp$, we are looking for the $\orderK$-least cl-models deductively
closed under support $k$-completion, \ie~models containing only the knowledge that 
can be inferred from $\lp$ and from the safe part of CWA identified by its k-maximal 
safe interpretation.

\begin{exmp}[running example cont.] \label{runex.4} Table~\ref{tab.4} extends
Table~\ref{tab.3}, by including models deductively closed under support k-completion. 
Note that now both $I_{8}$ and $I_{9}$
have been ruled out, as they are not minimal with respect to a given support,
\ie~$I_{8} \neq \min_{\orderK}(\supportClass{\supportphIx{I_{8}}}) =
\min_{\orderK}(\supportClass{\supportphIx{I_{5}}}) = KK(\lp \orK \supportphIx{I_{5}}) = I_{5}$
and $I_{9} \neq KK(\lp\orK \supportphIx{I_{9}}) = KK(\lp\orK \supportphIx{I_{6}}) = I_{6}$.
\begin{table}
\caption{Running example cont.: models deductively closed under support k-completion of $\lp$.} \label{tab.4}
\[
%{\small
{\footnotesize 
%\begin{array}{|c||ccc||ccc|c||c|c|c|c|c|} \hline
\begin{array}{ccccccccccccc} \hline\hline
&  \multicolumn{3}{c}{I_{i}} & \multicolumn{3}{c}{\supportphIx{I_{i}}} & &  &  &
\mbox{stable} & \mbox{supp.} & \mbox{deductively} \\ 
I_{i} \modelscl \lp & p & q & r & p & q & r & U_\lp(I_{i}) & KK(\lp) & WF(\lp) & \mbox{models} &
\mbox{models} & \mbox{closed models}\\ \hline
I_{1} & \botK & \botK & \botK & \botT & \botK & \botK &\{p\} &\bullet &         &  &  &    \\ %\hline
I_{2} & \botK & \topT & \botT & \botT & \botK & \botT &  \{p,r\} &       &         &   &  &   \\ %\hline
I_{3} & \botT & \botK & \botK & \botT & \botK & \botK &  \{p\}  &       & \bullet&\bullet  &\bullet&\bullet \\ %\hline
I_{4} & \botT & \botT & \topT & \botT & \botT & \botK & \{p,q\} &       &         &\bullet & \bullet&\bullet \\ %\hline
I_{5} & \botT & \topT & \botT & \botT & \botK & \botT & \{p,r\} &       &         &\bullet&\bullet&\bullet\\ %\hline
I_{6} & \botT & \topK & \topK & \botT & \botT & \botT & \{p,q,r\} &       &         & \bullet& \bullet&\bullet \\ %\hline
I_{7} & \topT & \topT & \botT & \botT & \botK & \botT & \{p,r\} &       &         &         & &\\ %\hline
I_{8} & \topK & \topT & \botT & \botT & \botK & \botT & \{p,r\}  &       &         &    & \bullet&  \\ %\hline
I_{9} & \topK & \topK & \topK & \botT & \botT & \botT &  \{p,q,r\} &       &         &    & \bullet& \\\hline\hline
\end{array}
}
\]
\end{table}
\okex
\end{exmp}

\nd Finally, we can note that an immediate consequence operator characterizing models deductively closed under support k-completion can be derived immediately from Theorem~\ref{thm.equivalence}, \ie~by relying on the operator $KK(\lp\orK\supportphIx{\cdot})$. In the following we present the
operator $\phiprimep$, which coincides with $KK(\lp\orK\supportphIx{\cdot})$, \ie~$\phiprimep(I) = KK(\lp \orK
\supportph)$ for any interpretation $I$, but does not require any, even intuitive, program
transformation like $\lp\orK\supportphIx{\cdot}$. 
 This may be important in the classical
logic programming case where $\lp\orK\supportphIx{\cdot}$ is not easy to define (as $\orK$
does not belong to the language of classical logic programs).
Therefore, the set of models deductively closed under support k-completion coincides with
the set of fixed-points of $\phiprimep$, which will be defined in terms of $\phip$ only.

Informally, given an interpretation $I$, $\phiprimep$ computes all the  knowledge that can be
inferred from the rules and the support of $\lp$ \wrt~$I$ without any other extra knowledge. Formally,

\begin{definition}[immediate consequence operator $\phiprimep$]\label{def.ewo}
Consider a logic program $\lp$ and an interpretation $I$.  The operator $\phiprimep$ maps
interpretations into interpretations and is defined as the limit of the sequence of
interpretations $\iterJI_{i}$ defined as follows: for any $i \geq 0$,
\[
\begin{array}{lcl}
\iterJI_{0} & = & \supportphIx{I} \ , \\ \\
\iterJI_{i+1} & = & \phip(\iterJI_{i}) \orK \iterJI_{i} \ .
\end{array}
\]
\okdef
\end{definition}

%here

\nd In the following, with $\iterJI_{i}$ we indicate the $i$-th
iteration of the immediate consequence operator $\phiprimep$, according to 
Definition~\ref{def.ewo}.

Essentially, given the current knowledge expressed by $I$ about an intended model of $\lp$,  
we compute first the support, $\supportphIx{I}$, and then cumulate all the implicit knowledge
that can be inferred from $\lp$, by starting from the support.

It is easy to note that the sequence $\iterJI_{i}$ is 
monotone non-decreasing under $\orderK$ and, thus has a limit.
The following theorem follows directly from Theorems~\ref{thm.phipmontone}
and~\ref{thm.supportmonotone}, and from the Knaster-Tarski theorem.

\begin{theorem}
$\phiprimep$ is monotone \wrt~$\orderK$. Therefore, 
$\phiprimep$ has a least (and a greatest) fixed-point under $\orderK$.
\end{theorem}

\nd Finally, note that  

\begin{itemize}
\item by definition $\phiprimep(I) = \phip(\phiprimep(I)) \orK \phiprimep(I)$,
and thus $\phip(\phiprimep(I)) \orderK \phiprimep(I)$; and

\item for fixed-points of $\phiprimep$ we have 
that $I = \phip(I) \orK I$ and, thus, $\phip(I) \orderK I$.
\end{itemize}

\nd Before proving the last theorem of this section, we need the following
lemma.

\begin{lemma}\label{lem.XX1}
Let $\lp$ be a logic program and let $I$ and $K$ be interpretations. If $K \modelscl \lp \orK
\supportph$ then  $\phiprimep(I) \orderK K$. \ok
\end{lemma}
\begin{proof}
Assume $K \modelscl \lp \orK \supportph$, \ie~by Lemma~\ref{lem.A}, $K = \phipx{\lp \orK
\supportph}(K) = \phip(K) \orK \supportph$. Therefore, $\supportph \orderK K$.
We show by induction on $i$ that $\iterJI_{i} \orderK K$ and, thus, at the 
limit $\phiprimep(I)\orderK K$.

\ii{i} Case $i=0$. By definition, $\iterJI_{0} = \supportphIx{I} \orderK K$.

\ii{ii} Induction step: suppose $\iterJI_{i} \orderK K$. Then by assumption 
and by induction we have that $\iterJI_{i+1}  =  \phip(\iterJI_{i}) \orK
\iterJI_{i} \orderK \phip(K) \orK K = \phip(K) \orK \phip(K) \orK \supportph = \phip(K) \orK
\supportph = K$, which concludes.\okproof
\end{proof}

\nd The following concluding theorem characterizes the set of models deductively 
closed under support k-completion in terms of fixed-points of $\phiprimep$. 

\begin{theorem} \label{lem.phiprime}
Let $\lp$ and $I$ be a logic program and an interpretation, respectively. Then
$\phiprimep(I) = KK(\lp \orK \supportp(I))$. 
\end{theorem}
\begin{proof}
The Kripke-Kleene model (for ease denoted $K$) of $\lp \orK \supportp(I)$ under $\orderK$, 
is the limit of the sequence 
\[
\begin{array}{lcl}
K_{0} & = & \IbotK \ , \\ \\
K_{i+1} & = & \phipx{\lp \orK \supportphIx{I}}(K_{i}) \ .
\end{array}
\]

\nd As $K \modelscl \lp \orK \supportphIx{I}$, by Lemma~\ref{lem.XX1}, $\phiprimep(I) \orderK
K$. Now we show that $K \orderK \phiprimep(I)$, by proving by induction on $i$ that
$K_{i} \orderK \phiprimep(I)$ and, thus, at the limit $K\orderK\phiprimep(I)$.

\ii{i} Case $i=0$. We have $K_{0}  = \IbotK \orderK \phiprimep(I)$. 

\ii{ii} Induction step: suppose $K_{i} \orderK \phiprimep(I)$. 
Then, by induction we have $K_{i+1}  =  \phipx{\lp \orK \supportphIx{I}}(K_{i})
\orderK \phipx{\lp \orK \supportphIx{I}} (\phiprimep(I))$.
As $\supportphIx{I} \orderK \phiprimep(I)$, by Lemma~\ref{lem.A} it follows that $K_{i+1} 
\orderK \phipx{\lp \orK \supportphIx{I}} (\phiprimep(I)) 
= \phip(\phiprimep(I)) \orK \supportphIx{I} \orderK
\phip(\phiprimep(I)) \orK \phiprimep(I) = \phiprimep(I)$, which concludes. \okproof
\end{proof}

\nd It follows immediately that 

\begin{corollary} \label{thm.pimodel}
An interpretation $I$ is a model deductively closed under support k-completion of $\lp$ iff $I$ is a fixed-point of $\phiprimep$.\ok
\end{corollary}

%-----------------------------------
%\subsection{Stable models} \label{results}
%-----------------------------------

We will now state that \emph{the set of stable models coincides with the set of models deductively closed under support k-completion}.  This statement implies that our approach leads to an epistemic characterization of the family of stable models. It also evidences the role of CWA in logic programming. Indeed, CWA can be seen as  the additional support of falsehood to be added cumulatively to the Kripke-Kleene semantics
to define some more informative semantics: the well-founded  and the stable model semantics.
Moreover, it gives a new fixed-point characterization of that family.
Our fixed-point characterization is based on $\phip$ only and neither requires any program
transformation nor separation of positive and negative literals/information. The proof of the
following stable model characterization theorem can be found in the appendix.

\begin{theorem}[stable model characterization] \label{theorem.sms}
Let $\lp$ and $I$ be a logic program and an interpretation, respectively.
The following statements are equivalent:
\begin{enumerate}
\item $I$ is a stable model of $\lp$;
\item $I$ is a  model deductively closed under support k-completion of $\lp$,\\ \ie~$I = \min_{\orderK}(\supportClass{\supportph})$;
\item $I = \phiprimep(I)$;
\item $I=  KK(\lp \orK \supportp(I))$. \ok
\end{enumerate}
\end{theorem}

\nd Considering a classical logic program $\lp$, a partial interpretation is a stable model of $\lp$ if and only if it is deductively closed under its greatest unfounded set completion, \ie~if and only if it coincides with the limit of the sequence:
\[
\begin{array}{lcl}
\iterJI_{0} & = & \neg.U_\lp(I) \ , \\ \\
\iterJI_{i+1} & = & \phip(\iterJI_{i}) \cup \iterJI_{i} \ .
\end{array}
\]

\nd Finally it is well-known that the least stable model
of $\lp$ \wrt~$\orderK$ coincides with $\lp$'s well-founded semantics. 
Therefore, our approach also provides new characterizations of the well-founded semantics
of logic programs over bilattices. Together with Theorem~\ref{thm.supmod.char}, we
have

\begin{corollary} \label{coro.wfs}
Let $\lp$ be a logic program. The following statements are equivalent:
\begin{enumerate}
\item $I$ is the well-founded semantics of $\lp$;
\item $I$ is the $\orderK$-least supported model of $\lp$, \ie~the $\orderK$-least  fixed-point of $\ecopB$;
\item$I$ is the $\orderK$-least model deductively closed under support k-completion of $\lp$,   \ie~the $\orderK$-least fixed-point of $\phiprimep$. \ok
\end{enumerate}
\end{corollary}

\nd Therefore, the well-founded semantics can  be characterized by means of the
notion of supported models only. Additionally, we now also know why $\ecopB$ characterizes
the well-founded model, while fails in characterizing stable models. Indeed, from $I = \ecopB(I)$
it follows that $I$ is a model of $\lp\orK\supportph$, which does not guarantee that $I$ is the
$\orderK$-least cl-model of $\lp\orK\supportph$ (see Example~\ref{runex.4}). Thus, $I$ does not
satisfy Theorem~\ref{thm.equivalence}.  If $I$ is the  $\orderK$-least fixed-point of $\ecopB$, then $I$ is both a cl-model of $\lp\orK\supportph$ and $\orderK$-least. Therefore, the
$\orderK$-least supported model is always a model deductively closed under support k-completion 
as well and, thus a stable model.

The following concluding example shows the various ways of computing the well-founded
semantics, according to the operators discussed in this study: $\psiprimep$ and
$\phiprimep$. But, rather than relying on $\cal FOUR$ as truth space, as we did in our running example,
we consider the bilattice of intervals over the unit $[0,1]$, used frequently for reasoning
under uncertainty.

\begin{exmp} \label{runex.6}
Let us consider  the bilattice of intervals
\mbox{$\tuple{[0,1] \times [0,1],\orderT,\orderK}$} introduced
in Example~\ref{exmy}.
Consider the following logic program $\lp$,
\[
\begin{array}{lcl}
A & \leftarrow & A \orT B \\
B & \leftarrow & (\negT C \andT A) \orT \tuple{0.3, 0.5} \\ 
C & \leftarrow & \negT B \orT \tuple{0.2, 0.4} \\
\end{array}
\]

\nd The table below shows the computation of the Kripke-Kleene semantics of $\lp$,
$KK(\lp)$, as $\orderK$-least fixed-point of $\phip$.
\[
\begin{array}{cccc} \hline\hline
A & B & C & K_{i} \\ \hline
\tuple{0,1} & \tuple{0,1} & \tuple{0,1} & K_{0} \\ %\hline
\tuple{0,1} & \tuple{0.3,1} & \tuple{0.2,1} & K_{1} \\ %\hline
\tuple{0.3,1} & \tuple{0.3,0.8} & \tuple{0.2,0.7} & K_{2} \\ %\hline
\tuple{0.3,1} & \tuple{0.3,0.8} & \tuple{0.2,0.7} & K_{3} = K_{2} = KK(\lp) \\ \hline\hline
\end{array}
\]
\nd Note that knowledge increases during the computation as the intervals become more
precise, \ie~$K_{i} \orderK K_{i+1}$.

The following table shows us the computation of the well-founded semantics of $\lp$,
$WF(\lp)$, as $\orderK$-least fixed-point of $\psiprimep$.
{\small
\[
\begin{array}{cccccccc} \hline\hline
v^{W_j}_i& A & B & C & A & B & C & W_j \\ \hline \hline
v^{W_0}_0 &\tuple{0,0}&\tuple{0,0}&\tuple{0,0}&\tuple{0,1}&\tuple{0,1}&\tuple{0,1}& W_0 \\
v^{W_0}_1 &\tuple{0,0}&\tuple{0.3,0.5}&\tuple{0,1} &&&& \\
v^{W_0}_2 &\tuple{0.3,0.5}&\tuple{0.3,0.5}&\tuple{0,1} &&&& \\
v^{W_0}_3 &\tuple{0.3,0.5}&\tuple{0.3,0.5}&\tuple{0,1} &&&& \\ \hline \hline 
v^{W_1}_0 &\tuple{0,0}&\tuple{0,0}&\tuple{0,0}&\tuple{0.3,0.5}&\tuple{0.3,0.5}&\tuple{0,1}& W_1\\ 
v^{W_1}_1 &\tuple{0,0}&\tuple{0.3,0.5}&\tuple{0.5,0.7}&&&& \\
v^{W_1}_2 &\tuple{0.3,0.5}&\tuple{0.3,0.5}&\tuple{0.5,0.7}&&&& \\ 
v^{W_1}_3 &\tuple{0.3,0.5}&\tuple{0.3,0.5}&\tuple{0.5,0.7}&&&& \\ \hline  \hline
v^{W_2}_0 &\tuple{0,0}&\tuple{0,0}&\tuple{0,0}&\tuple{0.3,0.5}&\tuple{0.3,0.5}&\tuple{0.5,0.7}& W_2 \\
v^{W_2}_1 &\tuple{0,0}&\tuple{0.3,0.5}&\tuple{0.5,0.7}&&&& \\
v^{W_2}_2 &\tuple{0.3,0.5}&\tuple{0.3,0.5}&\tuple{0.5,0.7}&&&& \\ 
v^{W_2}_3 &\tuple{0.3,0.5}&\tuple{0.3,0.5}&\tuple{0.5,0.7}&&&& \\  \hline \hline
 &&&&\tuple{0.3,0.5}&\tuple{0.3,0.5}&\tuple{0.5,0.7}& W_3 = W_{2} = WF(\lp) \\ \hline\hline
\end{array}
\]
}

\nd Note that $W_{i} \orderK W_{i+1}$ and $KK(\lp) \orderK WF(\lp)$, as expected.
We conclude this example by showing the computation of the well-founded semantics of $\lp$,
as $\orderK$-least fixed-point of $\phiprimep$.
{\small
\[
\begin{array}{cccccccc} \hline\hline
\hypohIx{I_n}_{i}& A & B & C & A & B & C & I_n/\iterJIx{I_{n}}_{j} \\ \hline \hline
\hypohIx{I_0}_0 &\tuple{0,0}&\tuple{0,0}&\tuple{0,0}&\tuple{0,1}&\tuple{0,1}&\tuple{0,1}& I_0 \\
\hypohIx{I_0}_1 &\tuple{0,0}&\tuple{0,0.5}&\tuple{0,1} &&&& \\
\hypohIx{I_0}_2 &\tuple{0,0.5}&\tuple{0,0.5}&\tuple{0,1} &&&& \\
\hypohIx{I_0}_3 &\tuple{0,0.5}&\tuple{0,0.5}&\tuple{0,1} &&&&\\\hline\hline
&&&&\tuple{0,0.5}&\tuple{0,0.5}&\tuple{0,1}&\iterJIx{I_{0}}_{0} =\supportphIx{I_{0}}\\
&&&&\tuple{0,0.5}&\tuple{0.3,0.5}&\tuple{0.5,1}&\iterJIx{I_{0}}_{1} \\
&&&&\tuple{0.3,0.5}&\tuple{0.3,0.5}&\tuple{0.5,0.7}&\iterJIx{I_{0}}_{2} \\
&&&&\tuple{0.3,0.5}&\tuple{0.3,0.5}&\tuple{0.5,0.7}&\iterJIx{I_{0}}_{3} \\\hline\hline 
\hypohIx{I_1}_0&\tuple{0,0}&\tuple{0,0}&\tuple{0,0}&\tuple{0.3,0.5}&\tuple{0.3,0.5}&\tuple{0.5,0.7}&I_1\\ 
\hypohIx{I_2}_1 &\tuple{0,0}&\tuple{0,0.5}&\tuple{0,0.7}&&&& \\
\hypohIx{I_2}_2 &\tuple{0,0.5}&\tuple{0,0.5}&\tuple{0,0.7}&&&& \\ 
\hypohIx{I_2}_3 &\tuple{0,0.5}&\tuple{0,0.5}&\tuple{0,0.7}
&&&&  \\ \hline  \hline
&&&&\tuple{0,0.5}&\tuple{0,0.5}&\tuple{0,0.7}& \iterJIx{I_{1}}_{0} =\supportphIx{I_{1}} \\
&&&&\tuple{0,0.5}&\tuple{0.3,0.5}&\tuple{0.5,0.7}& \iterJIx{I_{1}}_{1} \\
&&&&\tuple{0.3,0.5}&\tuple{0.3,0.5}&\tuple{0.5,0.7}& \iterJIx{I_{1}}_{2} \\
&&&&\tuple{0.3,0.5}&\tuple{0.3,0.5}&\tuple{0.5,0.7}& \iterJIx{I_{1}}_{3} \\\hline \hline
&&&&\tuple{0.3,0.5}&\tuple{0.3,0.5}&\tuple{0.5,0.7}& I_2 = I_{1} = WF(\lp) \\ \hline\hline
\end{array}
\]
}

\nd Note how the knowledge about falsehood increases as our approximation to the intended model
increases, \ie~$\supportphIx{I_{i}} \orderK \supportphIx{I_{i+1}}$, while the degree of
truth decreases ($\supportphIx{I_{i+1}} \orderT \supportphIx{I_{i}}$).
Furthermore, note that $WF(\lp) \modelscl \lp$ and $\supportphIx{WF(\lp)} \orderK WF(\lp)$, \ie~$WF(\lp)$ is a supported model of $\lp$,
compliant to Corollary~\ref{coro.wfs}.

\okex
\end{exmp}

%-----------------------------------
\section{Conclusions} \label{concl}
%-----------------------------------

\nd Stable model semantics has become a well-established and accepted approach to the
management of (non-monotonic) negation in logic programs. 
In this study we have presented an alternative formulation to the Gelfond-Lifschitz transformation,
which has widely been used to formulate stable model semantics. 
Our approach is purely based on algebraic and semantical aspects of informative monotone
operators over bilattices. In this sense, we talk about epistemological foundation of the
stable model semantics. Our considerations are based on the fact that we regard the
closed world assumption as an additional source of falsehood and identify with the
\emph{support} the amount/degree of falsehood carried on by the closed world assumption. The
support is the generalization of the notion of the greatest unfounded set for classical logic programs to
the context of bilattices. The support is then used to complete the well-known Kripke-Kleene
semantics of logic programs. In particular, 

\begin{enumerate}
\item with respect to well-founded semantics, we have generalized both the fixed-point characterization  of the well-founded semantics of \cite{vanGelder91} to bilattices (Point 2. in Table~\ref{tab.xx}) and its model-theoretic characterization (Point 3. in Table~\ref{tab.xx}, see \eg~\cite{Leone97}). 

\item concerning stable model semantics, we have shown that 

\[
I \in \stable{\lp} \mbox{ iff } I = \min_{\orderK}(\supportClass{\supportph}) = KK(\lp \orK \supportph) = \phiprimep(I) \ ,
\]

\nd while previously stable models have been characterized by $I \in \stable{\lp}$ iff $I = \min_{\orderT}(\mod{{\lp^{I}}})$.

\end{enumerate}

\begin{table}
\caption{Well-founded semantics characterization: from classical logic to bilattices.} \label{tab.xx}
{\footnotesize
\begin{center}
\begin{tabular}{llcc} \hline\hline
1. & \multicolumn{3}{c}{$I$ is the well-founded semantics of $\lp$} \\ \hline
   &  & Classical logic $\{\botT, \botK,\topT\}$ & {\bf Bilattices} \\ %\hline
 \\
2. & $\orderK$-least  $I$ s.t.  &  $I  = W_{\lp}(I) = T_\lp(I) 
\cup \neg.U_\lp(I)$ &  $\mathbf{I = \ecopB(I) = \phip(I) \orK \supportp(I)}$ \\ %\hline
\\
3. &  $\orderK$-least model $I$ s.t. & $\neg. U_\lp(I)\subseteq I$ & $\mathbf{\supportp(I) \orderK I}$ \\\hline\hline
\end{tabular}
\end{center}
}
\end{table}

\nd Our result indicates that the support may be seen as the added-value to the Kripke-Kleene semantics and evidences the role of CWA in the well-founded and stable model semantics. It also shows that a separation of positive and negative information is nor necessary (as required by the
Gelfond-Lifschitz transform), nor is any program transformation required. 

%We have also shown that the well-founded semantics can be
%characterized as the knowledge minimal interpretation containing its support, \ie~$WF(\lp)=
%\min_{\orderK}(\{I \colon \supportph \orderK I\})$.

As our approach is rather general and abstracts from the underlying logical formalism (in our
case logic programs), it may be applied to other contexts as well.

%---------------------
\appendix

%-----------------------------------
\section{Proof of  Theorem~\ref{theorem.sms}} \label{sec.app.sms}
%-----------------------------------

\renewcommand{\thesection}{\Alph{section}}
\renewcommand{\thelemma}{\thesection.\arabic{lemma}}

\nd This part is devoted to the proof of  Theorem~\ref{theorem.sms}. It relies on the following
intermediary results. We start by providing lemmas to show that fixed-points of $\phiprimep$ are
stable models.

\begin{lemma}\label{lem.l1+}
If $I \orderT J$ and $J \orderK I$, then $\IbotT \andK 
\psip(x,I) = \IbotT \andK \psip(x,J)$, for any interpretation $x$. \ok
\end{lemma}
\begin{proof}
Using the antimonotonicity of $\psip$ \wrt~$\orderT$ for its
second argument, we have $\IbotT \orderT \psip(x,J)
\orderT \psip(x,I)$. From Lemma~\ref{prop.y1}, we have
$ \IbotT  \andK  \psip(x,I)\orderK \psip(x,J)$. Using the interlacing
conditions, we have $ \IbotT  \andK  \psip(x,I) \orderK \IbotT  \andK 
\psip(x,J)$.
Now, using the monotonicity of $\psip$ \wrt~$\orderK$ and the
interlacing conditions, we have $\IbotT  \andK  \psip(x,J) \orderK
\IbotT  \andK  \psip(x,I)$. It results that
$\IbotT  \andK  \psip(x,I) = \IbotT  \andK  \psip(x,J)$.\okproof
\end{proof}

\nd Similarly, we have

\begin{lemma}\label{lem.l1++}
If $J \orderT I$ and $J \orderK I$, then $\IbotT \andK 
\psip(I,x) = \IbotT \andK \psip(J,x)$, for any interpretation $x$. \ok
\end{lemma}
\begin{proof}
Using the monotonicity of $\psip$ w.r.t. $\orderT$ for its
first argument, we have $\IbotT \orderT \psip(J,x)
\orderT \psip(I,x)$. From Lemma~\ref{prop.y1}, we have
$\IbotT  \andK  \psip(I,x) \orderK \psip(J,x)$. Using the interlacing
conditions, we have $\IbotT  \andK  \psip(I,x)  \orderK \IbotT  \andK 
\psip(J,x)$.
Now, using the monotonicity of $\psip$ \wrt~$\orderK$ and the
interlacing conditions, we have $\IbotT  \andK  \psip(J,x) \orderK
\IbotT  \andK  \psip(I,x)$. It results that $\IbotT  \andK 
\psip(I,x) = \IbotT  \andK  \psip(J,x)$.\okproof
\end{proof}

%---------------------------------------------

\begin{lemma}\label{lem.l40}
If $I=\phip(I)$ then $\hypohIx{I}_i \orderT \supportphIx{I} \orderT 
I$, for all $i$. \ok
\end{lemma}
\begin{proof}
By Theorem~\ref{thm.supportcomp},  the sequence 
$\hypohIx{I}_i$ is monotone non-decreasing under $\orderT$ and 
$\hypohIx{I}_i \orderT \supportphIx{I}$. Now, we show by induction on 
$i$ that $\hypohIx{I}_i \orderT I$ and, thus, at the limit 
$\supportphIx{I} \orderT I$.

\ii{i} Case $i=0$. $\hypohIx{I}_0 = \IbotT \orderT I$.

\ii{ii} Induction step: let us assume that $\hypohIx{I}_i \orderT I$ holds.
By Lemma~\ref{prop.p1}, $\hypohIx{I}_i \orderT \hypohIx{I}_i 
\oplus I \orderT I$ follows.
We also have $I \orderK \hypohIx{I}_i \oplus I$ and $\hypohIx{I}_i 
\orderK \hypohIx{I}_i \oplus I$.
It follows from Lemma~\ref{lem.l1+} and Lemma~\ref{lem.l1++} that
$\hypohIx{I}_{i+1} = \IbotT \otimes \psip(\hypohIx{I}_i \oplus I, 
\hypohIx{I}_i \oplus I)=
\IbotT \otimes \psip(\hypohIx{I}_i,I)$.
By induction $\hypohIx{I}_i \orderT I$, so from $I=\phip(I)$,
$\hypohIx{I}_{i+1} = \IbotT \otimes \psip(\hypohIx{I}_i, I) \orderT 
\psip(\hypohIx{I}_i,I)
\orderT \psip(I,I) = \phip(I) = I$ follows.\okproof
\end{proof}

\begin{lemma}\label{lem.l15A+}
If $I=\phip(I)$ then for any  $i$, $\supportpbotTIx{I} \orderK 
\hypohIx{I}_i \orderK v^{I}_i$ and, thus, at the limit 
$\supportpbotTIx{I} \orderK \psiprimep(I)$. \ok
\end{lemma}
\begin{proof}
By Theorem~\ref{thm.supportcomp}, $\supportpbotTIx{I} 
\orderK \hypohIx{I}_i $, for all $i$.
We know that $v^{I}_i$ 
converges to $\psiprimep(I)$. We 
show by induction on $i$ that $\hypohIx{I}_i \orderK v^{I}_i$. 
Therefore, at the limit $\supportpbotTIx{I} \orderK \psiprimep(I)$.

\ii{i} Case $i=0$. $\hypohIx{I}_0 = \IbotT  \orderK \IbotT = v^{I}_0$.

\ii{ii} Induction step: assume that $\hypohIx{I}_i \orderK v^{I}_i$.
By definition, $\hypohIx{I}_{i+1} = \IbotT \andK \phip(I \orK 
\hypohIx{I}_{i}) = \IbotT \andK \psip(I \orK \hypohIx{I}_{i}, I\orK 
\hypohIx{I}_{i})$. 
By Lemma~\ref{lem.l40}, $\hypohIx{I}_{i} \orderT 
I$.
By Lemma~\ref{prop.p1}, $\hypohIx{I}_i \orderT 
\hypohIx{I}_i \oplus I \orderT I$ follows.
We also have $I \orderK \hypohIx{I}_i \oplus I$ and $\hypohIx{I}_i 
\orderK \hypohIx{I}_i \oplus I$.
It follows from Lemma~\ref{lem.l1+} and Lemma~\ref{lem.l1++} that
$\hypohIx{I}_{i+1} = \IbotT \otimes \psip(\hypohIx{I}_i \oplus I, 
\hypohIx{I}_i \oplus I)=\IbotT \otimes \psip(\hypohIx{I}_i,I)$.
By the induction hypothesis we know that $\hypohIx{I}_i \orderK 
v^{I}_i$ for any $n$. Therefore,
$\hypohIx{I}_{i+1} \orderK \IbotT \andK \psip(v^{I}_i, I) \orderK 
\psip(v^{I}_i, I) = v^{I}_{i+1}$ follows, which concludes. \okproof
\end{proof}

\begin{lemma}\label{lem.s1}
Let $\lp$ and $I$ be a logic program and an interpretation, respectively. If $I$ is a supported model then $\supportph = \IbotT \andK I$.\ok
\end{lemma}
\begin{proof}
By Equation~\ref{eq.supporteq} and Theorem~\ref{thm.model2}, $\supportph = \IbotT \andK \phip(I \orK
\supportph) = \IbotT \andK I$. \okproof
\end{proof}

\begin{lemma}\label{lem.X2}
If $I = \phiprimep(I)$ then we have:
\begin{enumerate}
\item $\supportpbotTIx{I}  \orderT \psiprimep(I)  \orderT I$; and
\item $\supportpbotTIx{I}  \orderK \psiprimep(I)  \orderK I$.
\end{enumerate}
\end{lemma}
\begin{proof}
By Corollary~\ref{thm.pimodel} and by Lemma~\ref{lem.s1}, $\supportpbotTIx{I} = 
\IbotT \andK I$ and $I = \phip(I)$.
From Lemma~\ref{lem.l15A+}, 
$\supportpbotTIx{I}  \orderK \psiprimep(I)$.
By definition of 
$\psiprimep$, $\psiprimep(I) = \lfp_{\orderT} (\lambda
x.\psip(x,I))$. But, 
$I = \phip(I) = \psip(I,I)$, thus
$\psiprimep(I)  \orderT I$. 

Now we show by induction on $i$, that 
$\hypohIx{I}_i \orderT v^{I}_i$. Therefore, at the limit, 
$\supportpbotTIx{I} \orderT \psiprimep(I)$ and, thus, 
$\supportpbotTIx{I}  \orderT \psiprimep(I)  \orderT I$ hold. 

\ii{i} Case $i=0$. $\hypohIx{I}_0 = \IbotT \orderT \IbotT =  v^{I}_0$.

\ii{ii} Induction step: let us assume  that  $\hypohIx{I}_i \orderT 
v^{I}_i$ holds.
 From Lemma~\ref{lem.l40}, we have $\hypohIx{I}_i \orderT I$ and, 
thus, by Lemma~\ref{prop.p1},
$\hypohIx{I}_i \orderT \hypohIx{I}_i \oplus I \orderT I$ follows.
We also have $I \orderK \hypohIx{I}_i  \oplus I$ and $\hypohIx{I}_i 
\orderK \hypohIx{I}_i \oplus I$.
Then, from Lemma~\ref{lem.l1+} and Lemma~\ref{lem.l1++},
$\hypohIx{I}_{i+1} = \IbotT \otimes \psip(\hypohIx{I}_i \oplus I, 
\hypohIx{I}_i \oplus I) =
\IbotT \otimes \psip(\hypohIx{I}_i, I)$. By induction $\hypohIx{I}_i 
\orderT v^{I}_i$, so by Lemma~\ref{prop.p9} we have
$\hypohIx{I}_{i+1} = \IbotT \otimes \psip(\hypohIx{I}_i,I) \orderT 
\psip(\hypohIx{I}_i,I)
\orderT \psip(v^{I}_i,I) = v^{I}_{i+1}$, which concludes.

Finally, 
from $\supportpbotTIx{I}  \orderT \psiprimep(I)  \orderT I$ and by 
Lemma~\ref{prop.y1} 
we have $\psiprimep(I)  \orderK  I \orK 
\supportpbotTIx{I}   = I$, so 
$\supportpbotTIx{I}  \orderK 
\psiprimep(I)  \orderK I$. \okproof
\end{proof}

\nd Now we are ready to show that fixed-points of $\phiprimep$ are stable models.

\begin{theorem} \label{thm.TX2}
Every fixed-point  of $\phiprimep$ is a stable model of $\lp$. \ok
\end{theorem}
\begin{proof}
Assume $I = \phiprimep(I)$. Let us show that $I = 
\psiprimep(I)$. From Lemma~\ref{lem.X2}, we know that $\psiprimep(I) 
\orderK I$. Now, let us show by induction on $i$ that 
$\iterJI_i \orderK \psiprimep(I)$. Therefore, at the limit $I = \phiprimep(I) \orderK
\psiprimep(I)$ and, thus, $I = \psiprimep(I)$.

\ii{i} Case $i=0$. $\iterJI_0  = 
\supportpbotTIx{I}  \orderK \psiprimep(I)$, by Lemma~\ref{lem.X2}.

\ii{ii} Induction step: let us assume  that  $\iterJI_i \orderK 
\psiprimep(I)$ holds. By definition,

$\iterJI_{i+1} = \phip(\iterJI_{i}) \orK \iterJI_{i}$.
By induction 
$\iterJI_{i} \orderK \psiprimep(I)$. Therefore, $\iterJI_{i+1} 
\orderK \phip(\psiprimep(I)) \orK\psiprimep(I)$. But, by 
Lemma~\ref{lem.X2}, $\psiprimep(I) \orderK I$, 
so
$\phip(\psiprimep(I))$ = $\psip(\psiprimep(I)$, $\psiprimep(I)) 
\orderK$ $\psip(\psiprimep(I),I)$ = $\psiprimep(I)$. Therefore, 
$\iterJI_{i+1} \orderK \psiprimep(I)$. \okproof
\end{proof}

\nd The following lemmas are needed to show the converse, \ie~that stable models are fixed-points of
$\phiprimep$.

\begin{lemma}\label{lem.X3}
If $I = \psiprimep(I)$ then we have:
\begin{enumerate}
\item $\supportpbotTIx{I} \orderK I$;
\item $\phiprimep(I) \orderK I$;
\item $\phiprimep(I) \orderT I$.
\end{enumerate}
\end{lemma}
\begin{proof}
Assume $I = \psiprimep(I)$. By 
Theorem~\ref{thm.psiprime1}, $I = \phip(I)$.
By Lemma~\ref{lem.l15A+}, $\supportpbotTIx{I} \orderK 
\psiprimep(I) = I$, which completes Point~1..

Now, we show by 
induction on $i$ that,  $\iterJI_i  \orderK I$  and $\iterJI_i  \orderT 
I$ and, thus, at the limit $\phiprimep(I) \orderK I$ and 
$\phiprimep(I) \orderT I$ hold.

\ii{i} Case $i=0$. By Point~1., $\iterJI_0  = \supportpbotTIx{I}  \orderK I$, 
while $\iterJI_0   = \supportpbotTIx{I} \orderT I$, by 
Lemma~\ref{lem.l40}.

\ii{ii} Induction step: let us assume  that 
$\iterJI_i  \orderK I$  and $\iterJI_i  \orderT I$ hold.
By 
definition, $\iterJI_{i+1} = \phip(\iterJI_{i}) \orK \iterJI_{i}$.
By induction $\iterJI_i  \orderK I$, thus
$\iterJI_{i+1} 
\orderK \phip(I) \orK I = I \orK I = I$, which completes Point~2.
From $\iterJI_i \orderK I$, $\phip(\iterJI_i) \orderK \phip(I) =I$ follows.
By induction we have $\iterJI_i  \orderT I$, 
thus $\iterJI_{i+1} \orderT \phip(\iterJI_i) \orK I = I$, which 
completes Point~3. \okproof
\end{proof}

\begin{lemma}\label{lem.X4}
If $I = \psiprimep(I)$ then $I \orderT \phiprimep(I)$.
\end{lemma}
\begin{proof}
Assume $I = \psiprimep(I)$. By 
Theorem~\ref{thm.psiprime1}, $I = \phip(I)$.
By Lemma~\ref{lem.l40} 
and Lemma~\ref{lem.X3}, $\supportpbotTIx{I}  \orderK I$ and 
$\supportpbotTIx{I}  \orderT I$, so
 by Lemma~\ref{prop.sms4}, 
$\supportpbotTIx{I}= \supportpbotTIx{I}  \andK \IbotT = I \andK 
\IbotT$.

Now, we show by induction on $i$, that $v^{I}_i \orderT 
\phiprimep(I)$. Therefore, at the limit, $I = \psiprimep(I) \orderT 
\phiprimep(I)$. 

\ii{i} Case $i=0$. $ v^{I}_0 = \IbotT \orderT \phiprimep(I)$.

\ii{ii} Induction step: let us assume  that  $v^{I}_i \orderT 
\phiprimep(I)$ holds.
By definition and by the induction hypothesis, 
$v^{I}_{i+1} = \psip(v^{I}_i,I) \orderT \psip(\phiprimep(I),I)$.
By 
Lemma~\ref{lem.X3}, $\phiprimep(I) \orderT I$. Therefore, since 
$\psip$ is antitone in the second argument under $\orderT$, 
$v^{I}_{i+1}  \orderT \psip(\phiprimep(I),\phiprimep(I)) = 
\phip(\phiprimep(I))$. It follows that 
$v^{I}_{i} \orK v^{I}_{i+1} \orderT \phip(\phiprimep(I)) \orK 
\phiprimep(I) = \phiprimep(I)$.  By Lemma~\ref{prop.sms2}, (by 
assuming, $x = v^{I}_{i}, z = v^{I}_{i+1}, y = \phiprimep(I)$), 
$v^{I}_{i+1} \orderK \phiprimep(I) \orK \IbotT$ follows. 
By 
Lemma~\ref{lem.X3}, both $\phiprimep(I) \orderT I$ and $\phiprimep(I) 
\orderK I$ hold. Therefore, by Lemma~\ref{prop.sms4}, 
$\phiprimep(I) \andK \IbotT = I \andK \IbotT = \supportpbotTIx{I}$. 
 From Lemma~\ref{lem.l15A+}, $\phiprimep(I) \andK \IbotT = 
\supportpbotTIx{I} \orderK v^{I}_{i+1} \orderK \phiprimep(I) \orK 
\IbotT$. Therefore, by Lemma~\ref{prop.sms3}, it follows that 
$v^{I}_{i+1} \orderT \phiprimep(I)$, which concludes the proof. \okproof
\end{proof}

\nd We can now prove that every stable model is indeed a 
fixed-point of $\phiprimep$, which concludes the characterization of 
stable models on bilattices.

\begin{theorem} \label{thm.TX3}
Every stable model of $\lp$ is a fixed-point  of $\phiprimep$. 
\ok
\end{theorem}l
\begin{proof}
Assume $I = \psiprimep(I)$.
By 
Lemma~\ref{lem.X3}, $\phiprimep(I) \orderT I$, while by 
Lemma~\ref{lem.X4},
$I \orderT \phiprimep(I)$. So $I = 
\phiprimep(I)$. \okproof
\end{proof}

\nd Finally, Theorem~\ref{theorem.sms} flollows directly from
Theorems~\ref{thm.TX2}, \ref{thm.TX3}, \ref{thm.equivalence} and Corollary~\ref{thm.pimodel}.

%{\footnotesize
%\bibliographystyle{plain}
%\bibliography{mybiblio,Stracciasubmittedbib}

\begin{thebibliography}{10}

\bibitem{Alcantara02}
Jo{\~a}o Alcant{\^a}ra, Carlos~Viegas Dam{\'a}sio, and Lu{\'i}s~Moniz Pereira.
\newblock Paraconsistent logic programs.
\newblock In {\em Proc.\ of the 8th European Conference on Logics in Artificial
  Intelligence (JELIA-02)}, number 2424 in Lecture Notes in Computer Science,
  pages 345--356, Cosenza, Italy, 2002. Springer-Verlag.

\bibitem{Anderson75}
Alan~R. Anderson and Nuel~D. Belnap.
\newblock {\em Entailment - the logic of relevance and necessity}.
\newblock Princeton University Press, Princeton, NJ, 1975.

\bibitem{Arieli02}
Ofer Arieli.
\newblock Paraconsistent declarative semantics for extended logic programs.
\newblock {\em Annals of Mathematics and Artificial Intelligence},
  36(4):381--417, 2002.

\bibitem{Arieli96}
Ofer Arieli and Arnon Avron.
\newblock Reasoning with logical bilattices.
\newblock {\em Journal of Logic, Language and Information}, 5(1):25--63, 1996.

\bibitem{Arieli98}
Ofer Arieli and Arnon Avron.
\newblock The value of the four values.
\newblock {\em Artificial Intelligence Journal}, 102(1):97--141, 1998.

\bibitem{Avron96}
Avi Avron.
\newblock The structure of interlaced bilattices.
\newblock {\em Journal of Mathematical Structures in Computer Science},
  6:287--299, 1996.

\bibitem{Belnap77a}
Nuel~D. Belnap.
\newblock A useful four-valued logic.
\newblock In Gunnar Epstein and J.~Michael Dunn, editors, {\em Modern uses of
  multiple-valued logic}, pages 5--37. Reidel, Dordrecht, NL, 1977.

\bibitem{BlairH89}
H.~Blair and V.~S. Subrahmanian.
\newblock Paraconsistent logic programming.
\newblock {\em Theoretical Computer Science}, 68:135--154, 1989.

\bibitem{Clark78}
K.L. Clark.
\newblock Negation as failure.
\newblock In Herv{\'{e}} Gallaire and Jack Minker, editors, {\em Logic and data
  bases}, pages 293--322. Plenum Press, New York, NY, 1978.

\bibitem{Damasio98}
Carlos~Viegas Dam{\'a}sio and Lu{\'i}s~Moniz Pereira.
\newblock A survey of paraconsistent semantics for logic programs.
\newblock In D.~Gabbay and P.~Smets, editors, {\em Handbook of Defeasible
  Reasoning and Uncertainty Management Systems}, pages 241--320. Kluwer, 1998.

\bibitem{Damasio01}
Carlos~Viegas Dam{\'a}sio and Lu{\'i}s~Moniz Pereira.
\newblock Antitonic logic programs.
\newblock In {\em Proceedings of the 6th European Conference on logic
  programming and Nonmonotonic Reasoning (LPNMR-01)}, number 2173 in Lecture
  Notes in Computer Science. Springer-Verlag, 2001.

\bibitem{Denecker98}
Marc Denecker.
\newblock {T}he well-founded semantics is the principle of inductive
  definition.
\newblock In J.~Dix, L.~Farinos~del Cerro, and U.~Furbach, editors, {\em Logics
  in Artificial Intelligence, Proceedings of JELIA-98}, number 1489 in Lecture
  Notes in Artificial Intelligence, pages 1--16. Springer Verlag, 1998.

\bibitem{Denecker01a}
Marc Denecker, Maurice Bruynooghe, and Victor Marek.
\newblock Logic programming revisited: logic programs as inductive definitions.
\newblock {\em ACM Transactions on Computational Logic (TOCL)}, 2(4):623--654,
  2001.

\bibitem{Denecker99}
Marc Denecker, Victor~W. Marek, and Miros{\l}aw Truszczy{\'n}ski.
\newblock {A}pproximating operators, stable operators, well-founded fixpoints
  and applications in nonmonotonic reasoning.
\newblock In J.~Minker, editor, {\em NFS-workshop on Logic-based Artificial
  Intelligence}, pages 1--26, 1999.

\bibitem{Denecker02}
Marc Denecker, Victor~W. Marek, and Miros{\l}aw Truszczy{\'n}ski.
\newblock {U}ltimate approximations in nonmonotonic knowledge representation
  systems.
\newblock In D.~Fensel, F.~Giunchiglia, D.~McGuinness, and M.~Williams,
  editors, {\em Principles of Knowledge Representation and Reasoning:
  Proceedings of the 8th International Conference}, pages 177--188. Morgan
  Kaufmann, 2002.

\bibitem{Denecker03}
Marc Denecker, Victor~W. Marek, and Miros{\l}aw Truszczy{\'n}ski.
\newblock Uniform semantic treatment of default and autoepistemic logics.
\newblock {\em Artificial Intelligence Journal}, 143:79--122, 2003.

\bibitem{Dunn76}
J.~Michael Dunn.
\newblock Intuitive semantics for first-degree entailments and coupled trees.
\newblock {\em Philosophical Studies}, 29:149--168, 1976.

\bibitem{Dunn86}
J.~Michael Dunn.
\newblock Relevance logic and entailment.
\newblock In Dov~M. Gabbay and Franz Guenthner, editors, {\em Handbook of
  Philosophical Logic}, volume~3, pages 117--224. Reidel, Dordrecht, NL, 1986.

\bibitem{vanEmden76}
M.~H.~Van Emden and R.~A. Kowalski.
\newblock The semantics of predicate logic as a programming language.
\newblock {\em Journal of the ACM (JACM)}, 23(4):733--742, 1976.

\bibitem{Fitting93}
M.~C. Fitting.
\newblock The family of stable models.
\newblock {\em Journal of Logic Programming}, 17:197--225, 1993.

\bibitem{Fitting02}
M.~C. Fitting.
\newblock Fixpoint semantics for logic programming - a survey.
\newblock {\em Theoretical Computer Science}, 21(3):25--51, 2002.

\bibitem{Fitting85}
Melvin Fitting.
\newblock A {K}ripke-{K}leene-semantics for general logic programs.
\newblock {\em Journal of Logic Programming}, 2:295--312, 1985.

\bibitem{Fitting91}
Melvin Fitting.
\newblock Bilattices and the semantics of logic programming.
\newblock {\em Journal of Logic Programming}, 11:91--116, 1991.

\bibitem{Fitting92}
Melvin Fitting.
\newblock Kleene's logic, generalized.
\newblock {\em Journal of Logic and Computation}, 1(6):797--810, 1992.

\bibitem{Gelfond88}
Michael Gelfond and Vladimir Lifschitz.
\newblock The stable model semantics for logic programming.
\newblock In Robert~A. Kowalski and Kenneth Bowen, editors, {\em Proceedings of
  the 5th International Conference on Logic Programming}, pages 1070--1080,
  Cambridge, Massachusetts, 1988. The {MIT} Press.

\bibitem{Gelfond91}
Michael Gelfond and Vladimir Lifschitz.
\newblock Classical negation in logic programs and disjunctive databases.
\newblock {\em New Generation Computing}, 9(3/4):365--386, 1991.

\bibitem{Ginsberg88a}
Matthew~L. Ginsberg.
\newblock Multi-valued logics: a uniform approach to reasoning in artificial
  intelligence.
\newblock {\em Computational Intelligence}, 4:265--316, 1988.

\bibitem{Herre97}
Heinrich Herre and Gerd Wagner.
\newblock Stable models are generated by a stable chain.
\newblock {\em Journal of Logic Programming}, 30(2):165--177, 1997.

\bibitem{Kunen87}
Kenneth Kunen.
\newblock Negation in logic programming.
\newblock {\em Journal of Logic Programming}, 4(4):289--308, 1987.

\bibitem{Leone97}
Nicola Leone, Pasquale Rullo, and Francesco Scarcello.
\newblock Disjunctive stable models: Unfounded sets, fixpoint semantics, and
  computation.
\newblock {\em Information and Computation}, 135(2):69--112, 1997.

\bibitem{Levesque84a}
Hector~J. Levesque.
\newblock A logic of implicit and explicit belief.
\newblock In {\em Proc.\ of the 3th Nat.\ Conf.\ on Artificial Intelligence
  (AAAI-84)}, pages 198--202, Austin, TX, 1984.

\bibitem{Levesque88}
Hector~J. Levesque.
\newblock Logic and the complexity of reasoning.
\newblock {\em Journal of Philosophical Logic}, 17:355--389, 1988.

\bibitem{Lloyd87}
John~W. Lloyd.
\newblock {\em Foundations of Logic Programming}.
\newblock Springer, Heidelberg, RG, 1987.

\bibitem{Loyer02d}
Yann Loyer and Umberto Straccia.
\newblock Uncertainty and partial non-uniform assumptions in parametric
  deductive databases.
\newblock In {\em Proc.\ of the 8th European Conference on Logics in Artificial
  Intelligence (JELIA-02)}, number 2424 in Lecture Notes in Computer Science,
  pages 271--282, Cosenza, Italy, 2002. Springer-Verlag.

\bibitem{Loyer02b}
Yann Loyer and Umberto Straccia.
\newblock The well-founded semantics in normal logic programs with uncertainty.
\newblock In {\em Proc.\ of the 6th International Symposium on Functional and
  Logic Programming (FLOPS-2002)}, number 2441 in Lecture Notes in Computer
  Science, pages 152--166, Aizu, Japan, 2002. Springer-Verlag.

\bibitem{Loyer03c}
Yann Loyer and Umberto Straccia.
\newblock The approximate well-founded semantics for logic programs with
  uncertainty.
\newblock In {\em 28th International Symposium on Mathematical Foundations of
  Computer Science (MFCS-2003)}, number 2747 in Lecture Notes in Computer
  Science, pages 541--550, Bratislava, Slovak Republic, 2003. Springer-Verlag.

\bibitem{Loyer03d}
Yann Loyer and Umberto Straccia.
\newblock Default knowledge in logic programs with uncertainty.
\newblock In {\em Proc.~of the 19th Int.~Conf.~on Logic Programming (ICLP-03)},
  number 2916 in Lecture Notes in Computer Science, pages 466--480, Mumbai,
  India, 2003. Springer Verlag.

\bibitem{Loyer03a}
Yann Loyer and Umberto Straccia.
\newblock The well-founded semantics of logic programs over bilattices: an
  alternative characterisation.
\newblock Technical Report ISTI-2003-TR-05, Istituto di Scienza e Tecnologie
  dell'Informazione, Consiglio Nazionale delle Ricerche, Pisa, Italy, 2003.

\bibitem{LukasiewiczT01}
Thomas Lukasiewicz.
\newblock Fixpoint characterizations for many-valued disjunctive logic programs
  with probabilistic semantics.
\newblock In {\em In Proceedings of the 6th International Conference on Logic
  Programming and Nonmonotonic Reasoning (LPNMR-01)}, number 2173 in Lecture
  Notes in Artificial Intelligence, pages 336--350. Springer-Verlag, 2001.

\bibitem{Moore84}
Robert~C. Moore.
\newblock Possible-world semantics for autoepistemic logic.
\newblock In {\em Proceedings of the 1st International Workshop on Nonmonotonic
  Reasoning}, pages 344--354, New Paltz, NY, 1984.

\bibitem{Ng91}
Raymond Ng and V.S. Subrahmanian.
\newblock Stable model semantics for probabilistic deductive databases.
\newblock In Zbigniew~W. Ras and Maria Zemenkova, editors, {\em Proc.\ of the
  6th Int.\ Sym.\ on Methodologies for Intelligent Systems (ISMIS-91)}, number
  542 in Lecture Notes in Artificial Intelligence, pages 163--171.
  Springer-Verlag, 1991.

\bibitem{Przymusinski90a}
T.~C. Przymusinski.
\newblock Extended stable semantics for normal and disjunctive programs.
\newblock In D.~H.~D. Warren and P.~Szeredi, editors, {\em Proceedings of the
  7th International Conference on Logic Programming}, pages 459--477. MIT
  Press, 1990.

\bibitem{Przymusinski90b}
T.~C. Przymusinski.
\newblock Stationary semantics for disjunctive logic programs and deductive
  databases.
\newblock In S.~Debray and H.~Hermenegildo, editors, {\em Logic Programming,
  Proceedings of the 1990 North American Conference}, pages 40--59. MIT Press,
  1990.

\bibitem{Przymusinski90}
Teodor~C. Przymusinski.
\newblock The well-founded semantics coincides with the three-valued stable
  semantics.
\newblock {\em Fundamenta Informaticae}, 13(4):445--463, 1990.

\bibitem{Reiter78}
Raymond Reiter.
\newblock On closed world data bases.
\newblock In Herv{\'{e}} Gallaire and Jack Minker, editors, {\em Logic and data
  bases}, pages 55--76. Plenum Press, New York, NY, 1978.

\bibitem{Reiter80}
Raymond Reiter.
\newblock A logic for default reasoning.
\newblock {\em Artificial Intelligence}, 13:81--132, 1980.

\bibitem{Tarski55}
A.~Tarski.
\newblock A lattice-theoretical fixpoint theorem and its applications.
\newblock {\em Pacific Journal of Mathematics}, (5):285--309, 1955.

\bibitem{vanGelder89}
Allen van Gelder.
\newblock The alternating fixpoint of logic programs with negation.
\newblock In {\em Proc.\ of the 8th ACM SIGACT SIGMOD Sym.\ on Principles of
  Database Systems (PODS-89)}, pages 1--10, 1989.

\bibitem{vanGelder91}
Allen van Gelder, Kenneth~A. Ross, and John~S. Schlimpf.
\newblock The well-founded semantics for general logic programs.
\newblock {\em Journal of the ACM}, 38(3):620--650, 1991.

\end{thebibliography}

\end{document}